\documentclass{article}

\usepackage{hyperref}

\usepackage[accepted]{icml2024}

\usepackage{microtype}
\usepackage{graphicx}
\usepackage{amsmath}
\usepackage{amssymb}
\usepackage{booktabs}
\usepackage{makecell}
\usepackage{enumitem}
\usepackage{centernot}
\usepackage{amsfonts}
\usepackage[normalem]{ulem}
\usepackage{dsfont}
\usepackage{multirow}
\usepackage{cuted}
\usepackage{capt-of}
\usepackage{bbm}
\usepackage{balance}
\usepackage{subcaption}

\usepackage{xspace}
\usepackage{float}

\graphicspath{{figures/}}

\usepackage[capitalize,noabbrev]{cleveref}
\usepackage[textsize=tiny]{todonotes}

\newcommand{\modelname}{VideoPoet}

\newcommand{\videoframe}[1]{
\includegraphics[width=0.23\textwidth]{#1}}

\makeatletter
\DeclareRobustCommand\onedot{\futurelet\@let@token\@onedot}
\def\@onedot{\ifx\@let@token.\else.\null\fi\xspace}

\def\eg{\emph{e.g}\onedot} 
\def\ie{\emph{i.e}\onedot} 
 
\def\etc{\emph{etc}\onedot}

\makeatother

\crefname{figure}{Fig.}{Figs.}
\Crefname{figure}{Fig.}{Figs.}

\icmltitlerunning{\modelname{}: A Large Language Model for Zero-Shot Video Generation}

\begin{document}


\twocolumn[
\icmltitle{\modelname{}: A Large Language Model for Zero-Shot Video Generation}



\icmlsetsymbol{equal}{*}

\begin{icmlauthorlist}
\icmlauthor{Dan Kondratyuk}{equal,goog} \hfill
\icmlauthor{Lijun Yu}{equal,goog,cmu} \hfill
\icmlauthor{Xiuye Gu}{equal,goog} \hfill
\icmlauthor{José Lezama}{equal,goog} \hfill
\icmlauthor{Jonathan Huang}{equal,goog} \hfill
\icmlauthor{Grant Schindler}{goog} \hfill
\icmlauthor{Rachel Hornung}{goog} \hfill
\icmlauthor{Vighnesh Birodkar}{goog} \hfill
\icmlauthor{Jimmy Yan}{goog} \hfill
\icmlauthor{Ming-Chang Chiu}{goog} \hfill
\icmlauthor{Krishna Somandepalli}{goog} \hfill
\icmlauthor{Hassan Akbari}{goog} \hfill
\icmlauthor{Yair Alon}{goog} \hfill
\icmlauthor{Yong Cheng}{goog} \hfill
\icmlauthor{Josh Dillon}{goog} \hfill
\icmlauthor{Agrim Gupta}{goog} \hfill
\icmlauthor{Meera Hahn}{goog} \hfill
\icmlauthor{Anja Hauth}{goog} \hfill
\icmlauthor{David Hendon}{goog} \hfill
\icmlauthor{Alonso Martinez}{goog} \hfill
\icmlauthor{David Minnen}{goog} \hfill
\icmlauthor{Mikhail Sirotenko}{goog} \hfill
\icmlauthor{Kihyuk Sohn}{goog} \hfill
\icmlauthor{Xuan Yang}{goog} \hfill
\icmlauthor{Hartwig Adam}{goog} \hfill
\icmlauthor{Ming-Hsuan Yang}{goog} \hfill
\icmlauthor{Irfan Essa}{goog} \hfill
\icmlauthor{Huisheng Wang}{goog} \hfill
\icmlauthor{David A. Ross}{goog} \hfill
\icmlauthor{Bryan Seybold}{equal,goog} \hfill
\icmlauthor{Lu Jiang}{equal,goog,cmu}
\end{icmlauthorlist}

\icmlaffiliation{goog}{Google}
\icmlaffiliation{cmu}{Carnegie Mellon University}

\icmlcorrespondingauthor{Lijun Yu}{lijuny@google.com}
\icmlcorrespondingauthor{Jonathan Huang}{jonathanhuang@google.com}
\icmlcorrespondingauthor{David Ross}{dross@google.com}
\icmlcorrespondingauthor{Bryan Seybold}{seybold@google.com}
\icmlcorrespondingauthor{Lu Jiang}{roadjiang@gmail.com}

\icmlkeywords{Video Generation}

\vskip 0.3in
]



\printAffiliationsAndNotice{\icmlEqualContribution} 

\begin{abstract}
We present \modelname{}, a model for synthesizing high-quality videos from a large variety of conditioning signals.
\modelname{} employs a decoder-only transformer architecture that processes multimodal inputs -- including images, videos, text, and audio.
The training protocol follows that of Large Language Models (LLMs), consisting of two stages: pretraining and task-specific adaptation. During pretraining, \modelname{} incorporates a mixture of multimodal generative objectives within an autoregressive Transformer framework.
The pretrained LLM serves as a foundation that is adapted to a range of video generation tasks.
We present results demonstrating the model's state-of-the-art capabilities in zero-shot video generation, specifically highlighting the generation of high-fidelity motions. Project page: \url{https://sites.research.google/videopoet/}.
\end{abstract}




\begin{figure*}
\centering
\includegraphics[width=.99\textwidth,trim={0 0.6cm 0 0},clip]{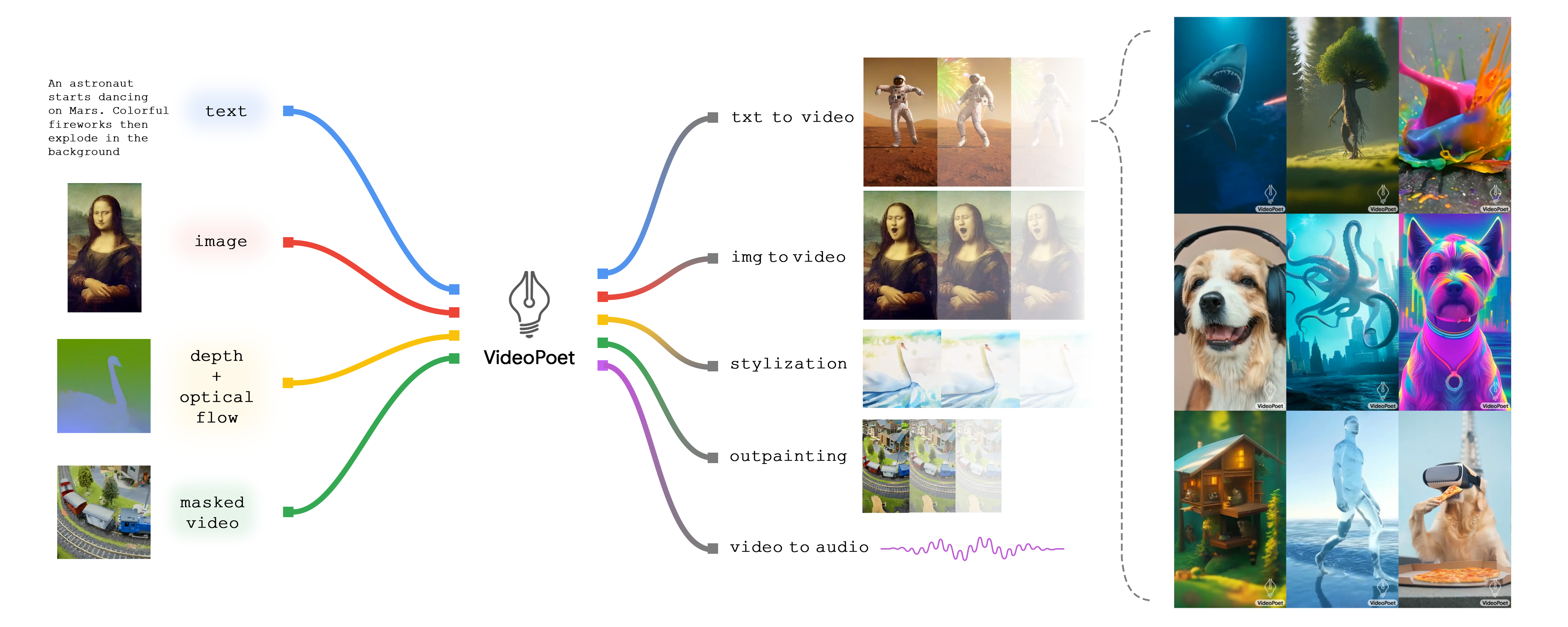}
\vspace{-2mm}
\caption{\textbf{\modelname{} Overview}: a versatile video generator that conditions on multiple types of inputs and performs a variety of video generation tasks.}
\label{fig:info_graphic}
\vspace{-4mm}
\end{figure*}

\vspace{-8mm}
\section{Introduction} 
\vspace{-1mm}
\label{sec:intro}

Recently, there has been a surge of generative video models capable of a variety of video creation tasks.
These include text-to-video~\cite{zhang2023show,singer2022make}, image-to-video~\cite{yu2023video}, video-to-video stylization~\cite{chen2023control,chai2023stablevideo,voleti2022mcvd}, and video editing~\cite{ceylan2023pix2video,wang2023zero,geyer2023tokenflow} among other video applications.
Most existing models employ diffusion-based methods for video generation. 
These video models typically start with a pretrained image model, such as Stable Diffusion~\cite{rombach2022high,podell2023sdxl}, that produces high-fidelity images for individual frames, and then fine-tune the model to improve temporal consistency across video frames. 

While Large Language Models (LLMs) are commonly used as foundation models across various modalities including language~\cite{brown2020language}, code~\cite{li2023starcoder,openai2023gpt4}, audio~\cite{rubenstein2023audiopalm}, speech~\cite{agostinelli2023musiclm}, and robotics~\cite{driess2023palm,brohan2023rt}, the diffusion model remains the predominant approach for video generation.
Although early research has demonstrated the effectiveness
of LLMs in text-to-image generation, \eg, DALL-E~\cite{ramesh2022hierarchical}, Parti~\cite{yu2022scaling} and~\cite{ding2021cogview}, and text-to-video, \eg, CogVideo~\cite{hong2022cogvideo}), language models have not reached a level of quality on par with video diffusion models in tasks like text-to-video generation as shown in previous studies~\cite{nash2022transframer,villegas2022phenaki}.
In contrast to training exclusively for text-to-video tasks, the generative model of LLMs in the language domain emphasizes a large pretraining stage to learn a foundation~\cite{bommasani2021opportunities} by examining pretraining tasks that extend beyond text-to-video generation.

A notable advantage of employing LLMs in video generation  
lies in the ease of integrating existing LLM frameworks. 
This integration allows for reusing LLM infrastructure and leverages the optimizations our community has developed over many years for LLMs, including optimizations in learning recipes for model scaling~\cite{brown2020language,chowdhery2022palm}, training and inference infrastructure~\cite{du2022glam}, hardware, among other advancements.
This couples with their flexibility in encoding many diverse tasks in the same model~\cite{raffel2020exploring},
which stands in contrast to most diffusion models where architectural changes and adapter modules are the dominant approach used to adapt the model to more diverse tasks~\cite{zhang2023adding}.

In this paper, we exploit language models for video generation, following the canonical training protocols of LLMs in the language domain.
We introduce \emph{\modelname{}}, a language model for video generation. \modelname{} employs a decoder-only LLM architecture~\cite{anil2023palm,openai2023gpt4} that admits image, video, and audio modalities as discrete tokens, each produced by their respective tokenizer.

The training process of \modelname{} consists of two stages: (1) pretraining and (2) task-adaptation.
During pretraining, \modelname{} incorporates a mixture of multimodal pretraining objectives within an autoregressive transformer framework.
After pretraining, the model functions as a versatile multi-task video generation model such as text-to-video, image-to-video, video editing and video-to-video stylization.
These capabilities are inherently integrated into a single LLM, rather than relying on a separate generative model controlled by text prompts~\citep{tang2023any}.
During subsequent task-adaptation, the pretrained model can be further fine-tuned either to enhance its generation quality on the training tasks or to perform new tasks.

Experiments show \modelname{}'s state-of-the-art capabilities in generating videos with large and high-fidelity motions. 
With the powerful capabilities of the transformer architecture, \modelname{} can be straightforwardly trained on a multi-task, multimodal generative objective, allowing for generating consistent and realistic motion driven by text or other prompts. 
Furthermore, \modelname{} can synthesize coherent long videos of up to 10 seconds by autoregressively extending the content, conditioned on the last second of the generated video. 
%

We also demonstrate that \modelname{} is capable of zero-shot video generation. We use the term ``zero-shot video generation'' as \modelname{} processes new text, image, or video inputs that diverge from the training data distribution.
Furthermore, \modelname{} handles new tasks not included in its training. For example, \modelname{} is able to perform new editing tasks by sequentially chaining tasks together. 

Our contribution is a proof of concept demonstrating an understudied approach to high-quality video generation with LLMs, distinct from the dominant diffusion-based methods.
Specifically, the main contributions include:
\begin{itemize}[nosep, leftmargin=*]
    \item A method for training a Large Language Model (LLM) specifically for video generation, utilizing tokenized data that incorporates both text-paired and unpaired videos.
    \item A video super-resolution method that increases spatial resolution within the latent token space using a bidirectional transformer with efficient windowed local attention.
    \item Evaluations and demonstrations to highlight  \modelname{}'s competitive and state-of-the-art performance, especially in generating realistic and interesting videos with motion.
\end{itemize}

\begin{figure*}[tp]
\centering
\includegraphics[width=\textwidth]{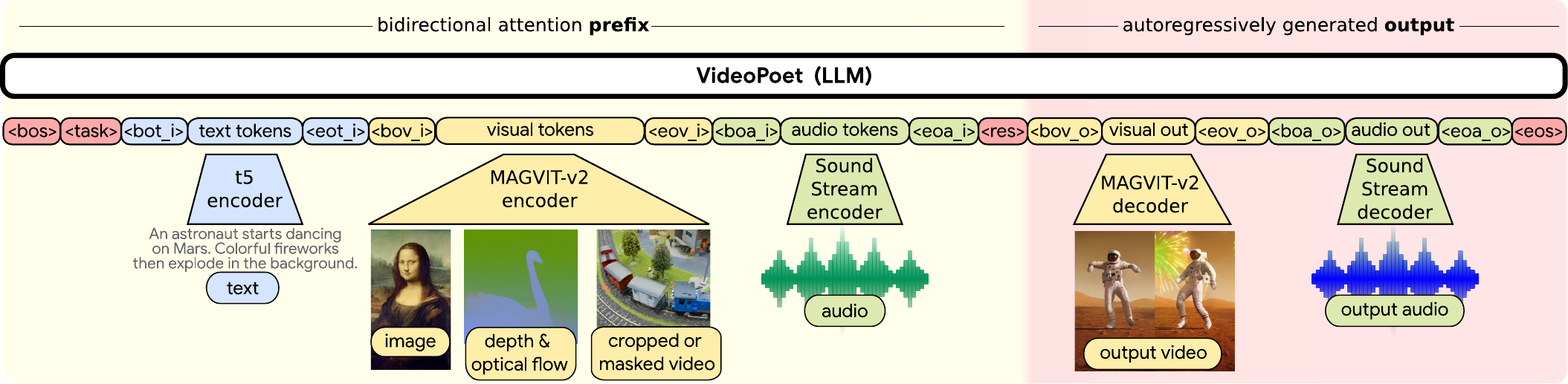}
\caption{
    \textbf{Sequence layout for \modelname{}}. We encode all modalities into the discrete token space, so that we can directly use large language model architectures for video generation.
    We denote special tokens in $<>$ (see Table~\ref{tab:special_tokens} for definitions).
    The modality agnostic tokens are in darker red; the text related components are in blue; the vision related components are in yellow; the audio related components are in green.
    The left portion of the layout on light yellow represents the bidirectional prefix inputs.
    The right portion on darker red represents the autoregressively generated outputs with causal attention.
}
\label{fig:vffm_schematic}
\vspace{-4mm}
\end{figure*}

\vspace{-2mm}
\section{Related Work} 
\label{sec:related_work}

\paragraph{Video diffusion models.}
Recently, numerous video generation methods use diffusion-based methods for text-to-video~\cite{ho2022imagen,blattmann2023align,zhang2023show,blattmann2023stable,he2023latent,zhou2022magicvideo,wang2023modelscope,ge2023preserve,wang2023internvid,wang2023videofactory,singer2022make,zhang2023show,zeng2023make} and video-to-video editing~\cite{liew2023magicedit,feng2023ccedit,esser2023structure,chen2023control}. 
As video diffusion models are usually derived from text-to-image diffusion models~\cite{ramesh2021zero,saharia2022photorealistic}, additional tasks and modalities are added via inference tricks~\cite{meng2021sdedit}, architectural changes~\cite{esser2023structure,liew2023magicedit} and adapter layers~\cite{zhang2023adding,guo2023animatediff}. 
Although these models are composable after training, they are not trained end-to-end in a unified framework.
Our multitask pretraining strategy in a single model improves performance and provides zero-shot video generation capabilities.

\vspace{-4mm}
\paragraph{Language models for video and image generation.}
Video language models are typically 
 derived from the general family of transformer-based language models~\cite{vaswani2017attention,raffel2020exploring} that easily combine multiple tasks in pretraining and demonstrate powerful zero-shot capabilities.
Image generation language models can generate images autoregressively~\cite{yu2022scaling} or via masked prediction~\cite{chang2022maskgit,chang2023muse}.
Both families have been extended to text-to-video~\cite{hong2022cogvideo,villegas2022phenaki,hu2023gaia,yan2021videogpt} using paired data.
While other text-to-video work with transformers only leverages video-text pairs for training, we also leverage unpaired videos (without text) and the same video for different tasks.
Since video language models can flexibly incorporate numerous tasks~\cite{yu2023magvit,nash2022transframer}, including video-to-video, we extend this family of work to text- and multimodal-conditioned tasks in this work with a synergistic pretraining strategy across various tasks.









\vspace{-4mm}
\paragraph{Pretraining task design in LLMs.}

As language models can easily incorporate multiple training tasks, task selection is an important area of research. 
GPT-3~\cite{brown2020language} and PaLM~\cite{chowdhery2022palm} demonstrate that training LLMs on diverse tasks leads to positive scaling effects on zero- and few-shot tasks.
Other approaches show that masking approaches are a valuable learning target~\cite{hoffmann2022training,yu2023magvit,yu2023language}. 
As the model size grows, training data must grow as well~\cite{hoffmann2022training} to maintain similar performance. 
Our pretraining strategy enables using the same video for multiple training tasks even without paired text. 
This design facilitates training on a large quantity of video-only examples, thereby decreasing the demand for video-text pairs.

\vspace{-4mm}
\section{Model Overview} \label{sec:method}
\vspace{-1mm}
We propose an effective method for video generation and related tasks from different input signals by leveraging large language models.
Our model consists of three components: (1) modality-specific tokenizers, (2) a language model backbone~(\Cref{fig:vffm_schematic}), and (3) a super-resolution module~(\Cref{fig:sr_diagram}).
The tokenizers map input data -- \ie image pixels, video frames, and audio waveforms -- into discrete tokens in a \emph{unified} vocabulary.
The visual and audio tokens are flattened into a sequence of integers.
Next, the LLM accepts these tokens as input along with text embeddings, and is responsible for generative multi-task and multimodal modeling. 
As illustrated in \Cref{fig:vffm_schematic}, \modelname{} conditions on text embeddings, visual tokens, and audio tokens, and autoregressively predicts visual and audio tokens.
Subsequently, the super-resolution module increases the resolution of the video outputs while refining visual details for higher quality.

\vspace{-3mm}
\subsection{Tokenization}
\label{sec:tokenization}
\vspace{-2mm}

We employ the MAGVIT-v2~\cite{yu2023language} tokenizer for joint image and video tokenization, and the  SoundStream~\cite{zeghidour2021soundstream} tokenizer for audio. 
Visual and audio vocabularies are concatenated into a unified vocabulary. 
The text modality is represented by embeddings. 

\vspace{-4mm}
\paragraph{Image and video tokenizer.}
Visual tokenizers are key to generating high-quality video content, often determining the upper limit of achievable video generation quality~\cite{yu2023language}.
Among existing tokenizers~\cite{esser2020taming,villegas2022phenaki,yu2023magvit,yu2023spae}, we choose the MAGVIT-v2~\cite{yu2023language} tokenizer due to its performance in visual quality and high compression capabilities, which effectively reduce the sequence length required by the LLM, thereby facilitating more efficient and effective learning.
%
%
Specifically, a video clip is encoded and quantized into an
integer sequence integers, with a decoder mapping back to  pixel space. MAGVIT-v2 tokenizes 17-frame 2.125-second 128$\times$128 resolution videos sampled at 8 fps to produce a latent shape of $(5, 16, 16)$, which is then flattened into 1280 tokens, with a vocabulary size of $2^{18}$. 
We also tokenize videos into portrait aspect ratio at 128$\times$224 resolution, producing a latent shape of $(5, 28, 16)$, or 2240 tokens.

\vspace{-1mm}
We enforce causal temporal dependency, which facilitates the generation of longer videos. 
To jointly represent images and videos, we encode the initial frame of a video or a static image into tokens with a consistent shape of $(1, 16, 16)$.
We use the COMMIT~\cite{yu2023magvit} encoding scheme to tokenize the inpainting and outpainting tasks. 

\vspace{-4mm}
\paragraph{Audio tokenizer.} We tokenize audio clips with a pretrained SoundStream~\cite{zeghidour2021soundstream} tokenizer.
We embed 2.125 seconds of audio to produce 106 latent frames with a residual vector quantizer (RVQ) of four levels.
Two possible choices exist for predicting the audio tokens with the RVQ representation: 1) sequentially predicting the entire audio clip from a lower to a higher RVQ level, or 2) simultaneously predicting all RVQ levels for a single audio token. Our results suggest that the former method demonstrates a slight advantage over the latter approach.
Finally, each RVQ level has a disjoint vocabulary with each level containing 1,024 codes.
This results in a combined audio vocabulary size of 4,096 codes.

\vspace{-4mm}
\paragraph{Text embedding as input.}
Pretrained text representations, in general, outperform training our model by learning text tokens from scratch.
We use pretrained language embeddings from a frozen T5 XL encoder~\cite{raffel2020exploring}. 
For tasks with text guidance, such as text-to-video, T5 XL embeddings are projected into the transformer's embedding space with a linear layer.



\vspace{-3mm}
\subsection{Language Model Backbone}
\vspace{-2mm}
After converting the image, video, and audio modalities into discrete tokens within a shared vocabulary, we can directly leverage a language model to generate videos and audios in the token space.
We use a prefix language model with a decoder-only architecture as the backbone.
By constructing different patterns of input tokens to output tokens during training, we can control the tasks the model is able to perform as explained in \Cref{sec:modeling}.

\vspace{-3mm}
\subsection{Super-Resolution} \label{sec:superresolution}
\vspace{-2mm}

\begin{figure}[t]
\centering
\includegraphics[width=\linewidth]{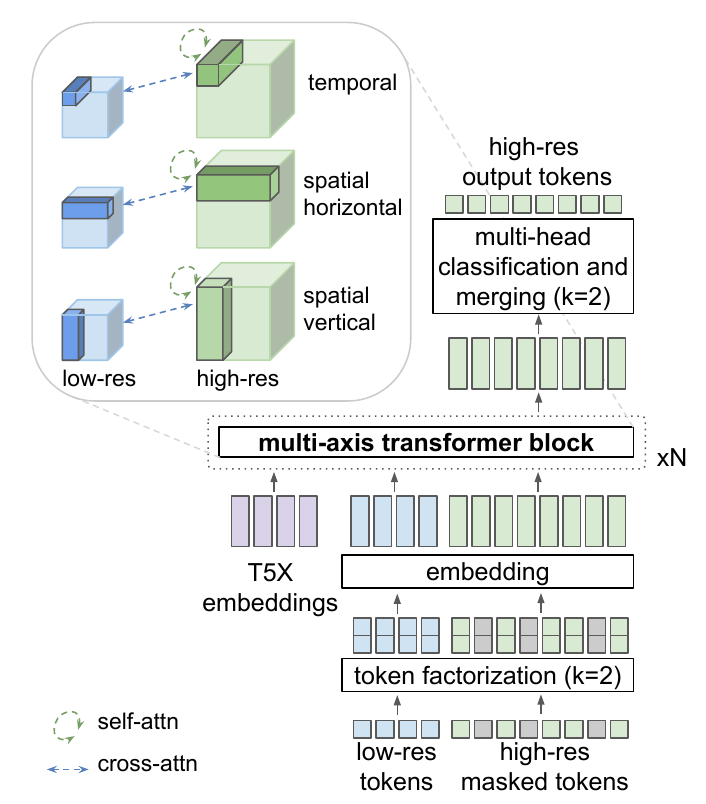}
\caption{\textbf{Custom transformer architecture for video super-resolution.}}
\label{fig:sr_diagram}
\vspace{-6mm}
\end{figure}

Generating high-resolution (HR) videos autoregressively entails heavy computational costs due to the increase in sequence length.
To illustrate this with an example, the video tokenizer of \Cref{sec:tokenization} operating on a $17\times896\times512$ video produces a sequence of $35,840$ tokens, making autoregressive sampling highly impractical.
Aiming at efficient and high-quality generative video upsampling, we develop a custom spatial super-resolution (SR) non-autoregressive video transformer~\citep{yu2023magvit} to operate in token space on top of the language model output.
To mitigate the computational requirements of the very long sequences involved, and in particular the quadratic memory of the self-attention layers, our design incorporates windowed local attention~\citep{gupta2022maskvit}.
Specifically, our SR transformer is composed of blocks of three transformer layers, each of which performs self-attention in a local window aligned with one of three axes~\citep{tu2022maxvit}: \emph{spatial vertical}, \emph{spatial horizontal} and \emph{temporal}.
The cross-attention layers attend to the low-resolution (LR) token sequence and are also divided into local windows, isomorphic to those of the self-attention layers.
All blocks also include cross-attention to T5 XL text embeddings. See~\Cref{fig:sr_diagram} for a schematic representation of the custom transformer architecture.

To account for the larger vocabulary size, we follow~\cite{yu2023language}, and train the SR transformer using token factorization with $k=2$ factors, which converts a $262,144$-way classification problem into two $512$-way classification problems. 
The LR token sequences are obtained by tokenizing bicubic-downsampled versions of the ground truth videos and applying noise augmentation~\citep{ho2022imagen} in the discrete latent space, to mitigate the distribution mismatch between real and generated videos.
Specifically, we randomly resample the value of a random subset of the LR tokens and independently drop the LR condition and text embeddings for 10\% of the training samples. During inference, we use non-autoregressive sampling \citep{chang2022maskgit,yu2023magvit} with classifier-free guidance independently on both the LR condition and the text embeddings \citep{brooks2023instructpix2pix}.
We use a cascade of two $2\times$ stages to generate videos of $896\times512$ resolution from the $224\times128$ base output of \modelname{}.
More implementaiton details can be found in the appendix. 

\vspace{-2mm}
\section{LLM Pretraining for Generation} 
\label{sec:modeling}
\vspace{-1mm}


\subsection{Task Prompt Design}
\label{sec:task_prompt_design}
\vspace{-2mm}

We design a pretraining task mixture, each with a defined prefix input and output. The model conditions on the prefix, applying the loss solely to the output.
\Cref{fig:vffm_schematic} shows a typical input-output sequence layout.
For each task, the input sequence may include three types of values: text embeddings (T5), visual tokens (MAGVIT-v2), and audio tokens (SoundStream). 
The model outputs two types of tokens: visual and audio tokens. To facilitate training, \modelname{} employs \emph{special tokens}, as listed in Appendix \Cref{tab:special_tokens}. 
In the following, we describe key designs for the task prompts.

\vspace{-4mm}
\paragraph{Pretraining tasks.}
We consider the following tasks. \emph{Unconditioned video generation}: Generate video frames without conditioning on an input. \emph{Text-to-video (T2V)}: Generate video from a text prompt. \emph{Video future prediction (FP)}: Given an input video of variable length, predict future frames. \emph{Image-to-video (I2V)}: Given the first frame of a video as an input image, predict the future frames. \emph{Video inpainting/outpainting (Painting)}: Given a masked video, predict the video with the masked contents filled in. \emph{Video stylization}: Given text, optical flow, and depth, predict the video frames (\Cref{sec:task_prompt_design}). \emph{Audio-to-video}: Given an input audio waveform, predict the corresponding video. \emph{Video-to-audio}: Given an input video, predict the corresponding audio waveform.
\emph{Audio-video continuation} (AVCont) given an input frame and its audio, predict the rest of the video and audio. In principle, the model can generate text, but we have not explicitly evaluated this ability.


To indicate the type of task, we condition on the \texttt{<task>} token, which has a unique value for each unique output.
We note that not all input variations need a new \texttt{<task>}; the model adapts to different context signals for identical outputs. 
For instance, text-to-video, image-to-video, and unconditioned video generation share the same \texttt{<task>}.
If a modality is absent in a task, related input/output tokens and special tokens are excluded, shortening the sequence.

\vspace{-3mm}
\paragraph{Representing an image as a video.}
In text-to-image pretraining, we omit the \texttt{<eos>} and \texttt{<eov\_o>} tokens from the input sequence, enabling continuous token generation for inference of longer videos.
This approach blurs the boundary between video and image generation tasks, enhancing cross-modality information sharing. 
This design leads to the prediction of higher-quality initial frames and reduces errors and artifacts in subsequent frames.

\vspace{-3mm}
\paragraph{Video token format.} 
We generate video tokens at two resolutions, 128$\times$128 and 128$\times$224, each available in two lengths: 17 frames and 41 frames, both encoded at 8 frames per second. 
Special conditioning tokens are used to signal the desired resolutions and durations for video generation.
Images are a special case of a 1-frame video, which we tokenize at 128$\times$128 resolution.

\vspace{-3mm}
\paragraph{Video stylization.} For video stylization, we adopt a method motivated  by~\cite{zhang2023adding,chen2023control,esser2023structure}, predicting videos from text, optical flow, and depth signals. The training task for stylization is to reconstruct the ground truth video from the given optical flow, depth, and text information, but
during inference, we apply optical flow and depth estimation on an input video but then vary the text prompt to generate a new style, \eg ``cartoon.''
%
%
Similar to~\cite{esser2023structure}, text dictates the output 
``content'' or appearance, while optical flow and depth guide its ``structure.''

\vspace{-2mm}
\subsection{Training Strategy} 
\label{sec:training}
\vspace{-2mm}

For multi-task training, we use the Alternating Gradient Descent (AGD) method~\cite{akbari2023alternating} to train videos of varying lengths. 
%
We design the tasks in the AGD format resulting in a near 0\% padding ratio, lower than that of the packing approach~\cite{raffel2020exploring}. 
This is accomplished by grouping tasks by sequence length and alternately sampling one group at each iteration.
%
Since sequence lengths are fixed and vary significantly across tasks, \eg, first frame and long video generation, we achieve efficient training with minimal padding.

We find that sampling from image and video datasets uniformly across time can lead to suboptimal results, as training on images can enhance the model's understanding of objects but does not capture any motions that are represented in video data.
Thus, we devise a two-stage pretraining strategy, where we augment our sampling weights to sample  image data 90\% of the time and video data 10\% of the time for the first 25\% iterations of training.
We then switch to training on video 90\% and image 10\% for the remaining iterations. 

We fine-tune our pretrained model for enhanced performance on specific tasks or for new task adaptation, such as text-to-video and image-to-video tasks, using a data subset of higher quality. These videos are sourced from broad internal sources with millions of videos that contain simpler clips, and are not manually tailored or selected. This results in improved generation quality, consistent with \citet{zhou2023lima}, and addresses decoding collapse issues, characterized by repetitive token predictions. 
Such fine-tuning not only diversifies outputs but also allows for a higher classifier-free guidance scale~\cite{ho2022classifier}, boosting overall quality. 


\vspace{-3mm}
\section{Experiments} \label{sec:experiments}

\vspace{-1mm}
\subsection{Experimental Setup}

\paragraph{Training tasks.}
We train the model on a mixture of pretraining tasks as detailed in~\Cref{sec:task_prompt_design}.
We finetune a model on a high-quality training subset for text-to-video evaluations, as discussed in \Cref{sec:training}.
Unless explicitly stated, we do not finetune on specific tasks for evaluations.

\vspace{-4mm}
\paragraph{Datasets.} 
\label{sec:exp_data}
We train on a total of 1B image-text pairs and $\sim$270M videos ($\sim$100M with paired text, of which $\sim$50M are used for high-quality finetuning,  and $\sim$170M with paired audio) from the public internet and other sources, \ie around 2 trillion tokens across all modalities. 
%
%
The data has been filtered to remove egregious content and sampled to improve contextual and demographic diversity.

\begin{table*}[tp]
\centering
\caption{\textbf{Pretraining task analysis on 300M models.} The top rows list models with 300M parameters, trained on a subset of the data, and are comparable to each other. The last row shows an 8B model trained on the entire dataset. \textbf{T2I} (text-to-image), \textbf{T2V} (text-to-video), \textbf{FP} (frame prediction), \textbf{Painting} (inpainting/outpainting), \textbf{Uncond} (unconditional generation), \textbf{AVCont} (audio-video continuation), and \textbf{SSL} (self-supervised learning).
}
\scriptsize
\begin{tabular}{l|cccccc|ccccc}
\toprule
\multicolumn{1}{c}{\multirow{4}{*}{Method}} & \multicolumn{6}{c|}{Pretraining Tasks}                                                                                                             & \multicolumn{5}{c}{Zero-shot Evaluation Benchmark}                                                                                                                                 \\ \cline{2-12} 
\multicolumn{1}{c}{}                        & \multirow{3}{*}{T2I} & \multirow{3}{*}{T2V} & \multirow{3}{*}{Uncond} & \multirow{3}{*}{FP} & \multirow{3}{*}{Painting} & \multirow{3}{*}{AVCont} & \multicolumn{2}{c}{T2V}                                 & \multicolumn{1}{c}{FP}   & \multicolumn{1}{c}{Inpainting} & \multicolumn{1}{c}{Outpainting}  \\ \cline{8-9}
\multicolumn{1}{c}{}                        &                      &                      &                         &                     &                           &                         & \multicolumn{1}{c}{MSR-VTT} & \multicolumn{1}{c}{UCF101} & \multicolumn{1}{c}{K600} & \multicolumn{1}{c}{SSv2}       & \multicolumn{1}{c}{SSv2}     \\
\multicolumn{1}{c}{}                        &                      &                      &                         &                     &                           &    & \multicolumn{1}{c}{CLIPSIM $\uparrow$}   & \multicolumn{1}{c}{FVD $\downarrow$}    & \multicolumn{1}{c}{FVD $\downarrow$}  & \multicolumn{1}{c}{FVD $\downarrow$}        & \multicolumn{1}{c}{FVD $\downarrow$}           \\
\midrule
T2V  && \checkmark&  &  &  & & 0.244&822&759&2,333&2,310                     \\
T2V+I & \checkmark & \checkmark &  &&&& 0.247&1,025&794&2,118&1,916                        \\
SSL & & & \checkmark & \checkmark & \checkmark & \checkmark & 0.226&1,742&700&1,093&1,500 \\
NO T2I &  & \checkmark & \checkmark & \checkmark & \checkmark & \checkmark & 0.235&1,008&755&95&389 \\
ALL & \checkmark & \checkmark& \checkmark & \checkmark & \checkmark & \checkmark& 0.240&1,085&729&127&636 \\ 
\midrule
ALL (8B) & \checkmark & \checkmark& \checkmark & \checkmark & \checkmark & \checkmark & 0.305	& 355	& 687	& 4.7 &	13.76  \\
\bottomrule
\end{tabular}
\label{tab:pretraining}
\vspace{-3mm}
\end{table*}

\vspace{-4mm}
\paragraph{Evaluation protocol.}
We employ a zero-shot generation evaluation protocol, 
as the model has not been trained on the training data of target benchmarks.
Specifically, the evaluation benchmark includes two text-to-video generation datasets, MSR-VTT~\cite{xu2016msr} and UCF-101~\cite{soomro2012ucf101}, as well as the frame prediction task on Kinetics 600 (K600)~\cite{carreira2018short}, in which the first 5 frames are provided as the condition to predict the next 11 frames. We also include inpainting and outpainting tasks~\cite{yu2023magvit} on Something-Something V2 (SSv2)~\cite{goyal2017something}. 

We employ widely used metrics such as Fréchet Video Distance (FVD)~\cite{unterthiner2018towards}, CLIP similarity score~\cite{wu2021godiva}, and Inception Score (IS)~\cite{saito2020train} for evaluation. 
Note that the specific metrics and evaluation methods vary across different datasets. Detailed information on these variations can be found in \Cref{sec:appendix-evaluation}. 
\textcolor{black}{
We include examples of the generated videos in the supplementary materials.}

\vspace{-2mm}
\subsection{Pretraining Task Analysis}
\vspace{-2mm}
We investigate the learning capabilities of different combinations of pretraining tasks using a model with 300 million parameters. 
All task combinations are trained using a learning rate of $10^{-3}$ for the same number of steps (300k) with a batch size of 1024.

For the analysis of pretraining tasks, we consider text-to-video (T2V), text-to-image (T2I), and four self-supervised learning (SSL) tasks: frame prediction (FP), central inpainting and central outpainting (Painting)~\cite{yu2023magvit} and audio-video continuation (AVCont) where the model is provided with the first frame and its corresponding audio to predict the subsequent 16 frames and matching audio. 
For each video task, we uniformly select 20\% of training samples from a random subset of 50 million videos.
For the text-to-image task, we randomly sample 50 million text-image pairs from our training dataset.
For tasks involving audio, our sampling is exclusive to videos that contain an audio track.

The evaluation results are presented in \Cref{tab:pretraining}.
We assess a model across the four tasks within the zero-shot evaluation benchmark: the T2V task on MSR-VTT~\cite{xu2016msr} and UCF 101~\cite{soomro2012ucf101}, the FP on K600~\cite{carreira2018short}, and central inpainting and outpainting on SSv2~\cite{goyal2017something}. The audio generation results (FAD) are reported in \Cref{fig:model_scale_b} of the Appendix.
In these experiments, we employ a single model to perform all the tasks. 
The model is not trained on the training data of these evaluation datasets, and thus it is a zero-shot evaluation.

The top rows of~\Cref{tab:pretraining} depict each pretraining task configuration of the 300 million parameter model, which are comparable in their setups.
Our evaluation benchmarks span diverse visual domains, posing a challenge to achieving consistent improvement across all of them.
Nevertheless, incorporating all pretraining tasks results in the best overall performance, on average, across all evaluated tasks. 
Additionally, the significant disparity observed in the ``SSL'' row suggests the limitations of self-supervised training and underscores the necessity for text-paired data during training. Both single-task and multi-task models are trained for the same number of steps. The minor decrease in performance of multi-task training in~\Cref{tab:pretraining} might be due to the insufficient training of each task.
The last row, ``ALL (8B)'', is the model with 8 billion parameters, trained on the pretraining tasks as discussed in \Cref{sec:method} and utilized significantly more compute. 

\begin{figure}[tp] 
\centering

\begin{subfigure}{\linewidth}  
\includegraphics[width=.95\linewidth]{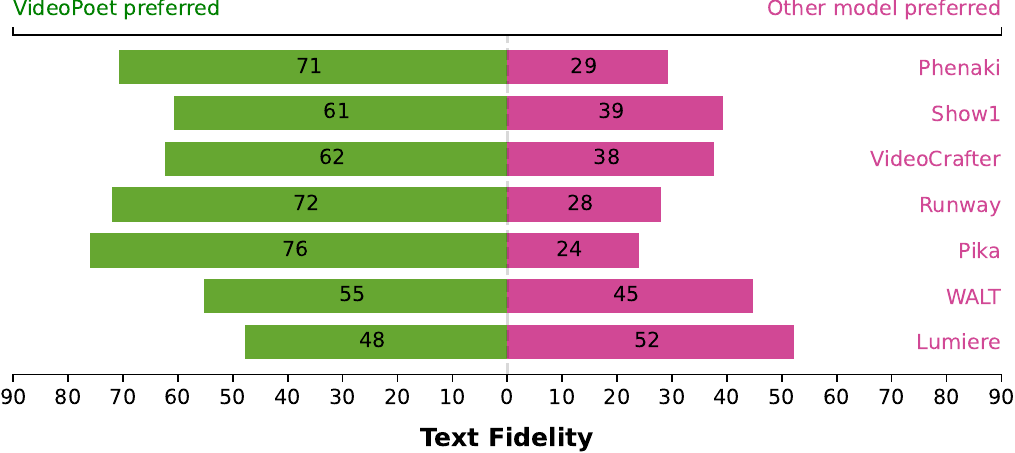}\vspace{3mm}
\end{subfigure}
\begin{subfigure}{\linewidth}  
\includegraphics[width=.95\linewidth]{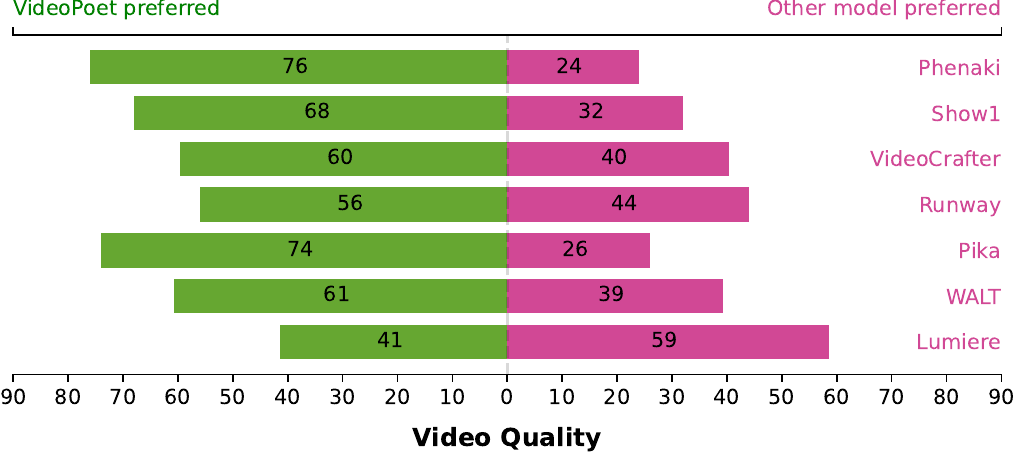}\vspace{3mm}
\end{subfigure}
\begin{subfigure}{\linewidth}  
\includegraphics[width=.95\linewidth]{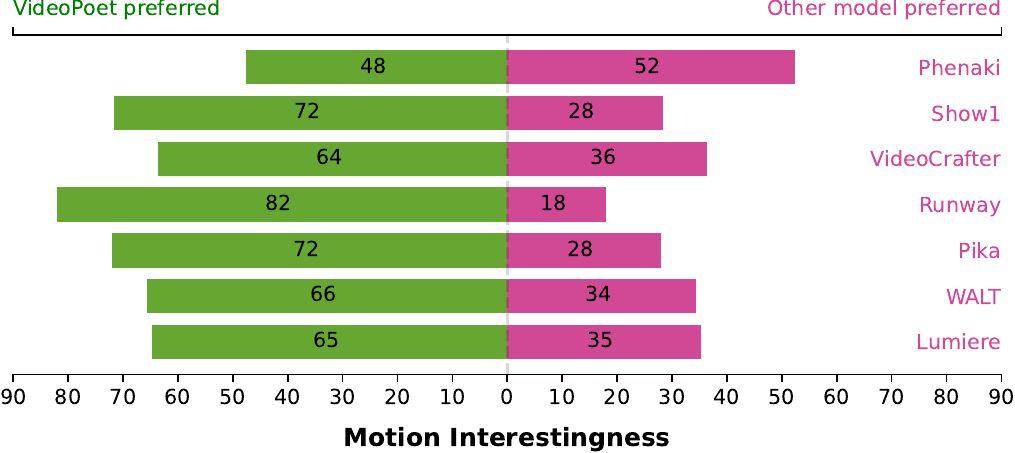}\vspace{3mm}
\end{subfigure}
\begin{subfigure}{\linewidth}  
\includegraphics[width=.95\linewidth]{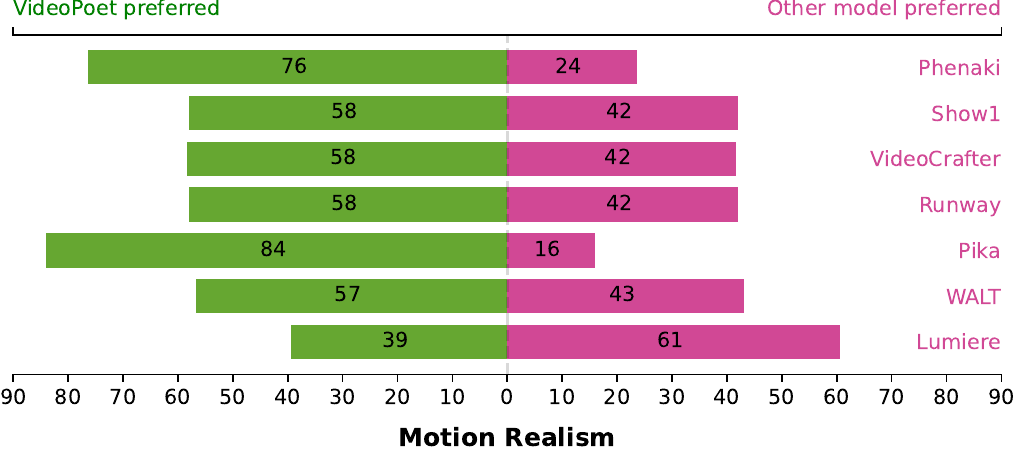}\vspace{3mm}
\end{subfigure}
\begin{subfigure}{\linewidth}  
\includegraphics[width=.95\linewidth]{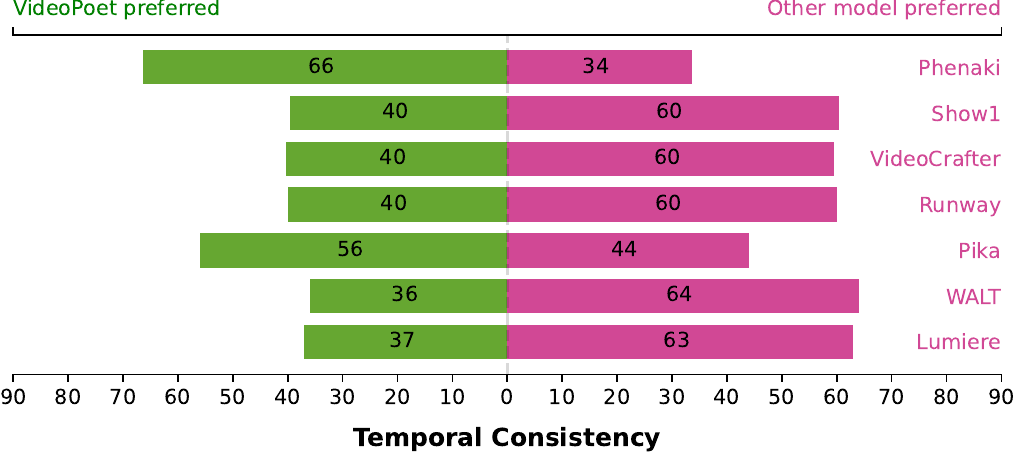}\vspace{3mm}
\end{subfigure}





\vspace{3mm}
\caption{
\textbf{Human evaluation results on text-to-video (T2V) generation.}
Green and pink bars represent the proportion of trials where \modelname{} was \textit{preferred over} or \textit{less preferred} to an alternative, respectively.
}
\label{fig:human_t2v_evals}
\end{figure}

\vspace{-2mm}
\subsection{Comparison with the State-of-the-Art}
\begin{table}[tp]
\centering
\caption{\textbf{Comparison on zero-shot text-to-video benchmarks.} See Appendix~\ref{sec:appendix-evaluation} for evaluation details.}
\resizebox{\linewidth}{!}{%
\begin{tabular}{lcccc}
\toprule
      Model & \multicolumn{2}{c}{MSR-VTT} & \multicolumn{2}{c}{UCF-101} \\
      & \multicolumn{1}{c}{CLIPSIM} & \multicolumn{1}{c}{FVD} & \multicolumn{1}{c}{FVD} & \multicolumn{1}{c}{IS} \\
\midrule
CogVideo (EN)~\yrcite{hong2022cogvideo} & \multicolumn{1}{c}{0.2631} & 1294 & 702 & 25.27 \\
MagicVideo~\yrcite{zhou2022magicvideo} & \multicolumn{1}{c}{-} & 998 & 655 & - \\
Video LDM~\yrcite{blattmann2023align} & \multicolumn{1}{c}{0.2929} & - &  551 & 33.45 \\
ModelScopeT2V~\yrcite{wang2023modelscope} & \multicolumn{1}{c}{0.2930} & 550 & - & - \\
InternVid~\yrcite{wang2023internvid} & \multicolumn{1}{c}{0.2951} & - & 617 & 21.04 \\
VideoFactory~\yrcite{wang2023videofactory} & \multicolumn{1}{c}{0.3005} & - & 410 & - \\
Make-A-Video~\yrcite{singer2022make} & \multicolumn{1}{c}{0.3049} & - & 367 & 33.00 \\
Show-1~\yrcite{zhang2023show} & \multicolumn{1}{c}{0.3072} & 538 & 394 & 35.42 \\
\bf \modelname{} (Pretrain) & \multicolumn{1}{c}{0.3049}  & \bf 213 & \bf 355 & \bf 38.44\\
\bf \modelname{} (Task adapt) & \multicolumn{1}{c}{\bf 0.3123} & - & - & - \\
\bottomrule
\end{tabular}
}
\label{tab:t2v_sota}
\vspace{-2mm}
\end{table}
\paragraph{Text-to-Video (T2V).}
\label{sec:exp_t2v}
\Cref{tab:t2v_sota} shows zero-shot text-to-video evaluation results on the common MSR-VTT~\cite{xu2016msr} and UCF-101~\cite{soomro2012ucf101} datasets. 
Our model performs favorably in terms of CLIP similarity and FVD scores on MSR-VTT and UCF-101. 
The pretrained foundation model already achieves competitive performance on all metrics.
After finetuned on high-quality subset of text-video pairs, \modelname{} achieves even better CLIPSIM on MSR-VTT.
More details on the evaluation settings can be found in~\Cref{sec:appendix-evaluation}.

\vspace{-4mm}
\paragraph{Human Evaluations with Text-to-Video (T2V).}
We analyze \modelname{} using human raters and compare with other recent
models: Show-1~\cite{zhang2023show},
VideoCrafter~\cite{chen2023videocrafter1}, Phenaki~\cite{villegas2022phenaki},  Pika~\cite{pika2023},  Gen2~\cite{runway2023}, WALT~\cite{gupta2023photorealistic}
and Lumiere~\cite{bar2024lumiere}.
Show-1, VideoCrafter, Pika, Gen2, WALT 
and Lumiere are video diffusion models
while Phenaki is a token-based model using  masked token modeling~\cite{chang2022maskgit}.
We ran the most up-to-date model versions as of January 2024
and note that WALT and Lumiere were concurrently developed
and not available during initial submission of this paper.

We first develop a unified evaluation prompt bank consisting of $\sim 250$
selected prompts from a variety of categories and styles.
Our prompts are
sourced from published prompt sets 
(\eg, Show-1, Video LDM~\cite{blattmann2023align}).
We select the prompts prior to generating videos
and fix these choices after initial selection. 
We also select preferentially for prompts that contain an explicit mention of motion so that the evaluation would not be biased for models that generate high quality videos that are almost still (e.g., ``person jumping off of a chair'' over ``person standing on a chair''). 
Note that due to time constraints, 
our experiments for Pika and Gen2 were run on a subset of
50 prompts due to having to submit these manually via their web
interface.  These 50 prompts were pre-selected (before any evaluations 
were run) so as to be representative of the entire set.

For this user study we use the fine-tuned version of \modelname{} as discussed in \Cref{sec:training}.
As part of our system we also use a fixed
negative prompt for all inference calls
as well as lightweight prompt rewriting.
We then compare \modelname{} against alternative models
in side-by-side fashion for each prompt. Raters are shown videos generated by two models at a time (in randomized order so as to not bias raters). We refer readers to 
Appendix~\ref{sec:appendix-human-evaluation}
for more details of our methodology.

Not all methods generate videos at the same size, aspect ratio, 
or framerate, and we thus resize and resample
each video to a fixed area while maintaining its original aspect ratio as well as common framerate. 
Raters are then asked to compare the videos along 5 dimensions and
for each dimension to report which video is better. 
The 5 dimensions are: (1) text fidelity (which video follows the text prompt most faithfully), (2) video quality, (3) motion ``interestingness'', (4) motion realism, and (5) temporal consistency.
Raters are required to go through a collection of training examples for each of these 5 dimensions before they begin. 

Our findings are summarized in \Cref{fig:human_t2v_evals}, where
green and pink bars  represent the  proportion  of trials  where \modelname{} was \textit{preferred} 
or \textit{less preferred} over an alternative, respectively.
%
We observe that 
\modelname{} is competitive with state of the art video
diffusion models despite taking a dramatically different approach, even outperforming these baseline models
along most dimensions.  
More specifically,  
\modelname{} achieves significant wins along
motion interestingness and realism and
Lumiere~\cite{bar2024lumiere} which is diffusion based and concurrent to our work, is the only model that outperforms \modelname{} on
Video Quality.


\subsection{Runtime}

For a batch size of 4 videos and generating 17 frames at 8fps using TPUv5p (4 chips) accelerators, our base model runs in 34s, the detokenizer (converting tokens to pixels) requires 1.3s and super-resolution is 6.8s — thus, amortized run time is about 5 seconds per second of output video. Note that our model has not been optimized, and any acceleration techniques applicable to standard LLMs could be applied here as well.
\vspace{-2mm}
\subsection{LLM's Diverse Capabilities in Video Generation}
\label{sec:capabilities_whole}

This subsection presents several capabilities we discover from the pretrained \modelname{}, shedding light on the LLM's promising potential in video generation. By combining the flexibility of our autoregressive language model to perform diverse tasks such as extending video in time, inpainting, outpainting, and stylization, \modelname{} accomplishes multiple tasks in a unified model.

\vspace{-4mm}
\paragraph{Coherent long video generation and image-to-video.}

\begin{figure}[t]
\addtocounter{figure}{-1}

\begin{subfigure}{1.0\linewidth}
\centering
\videoframe{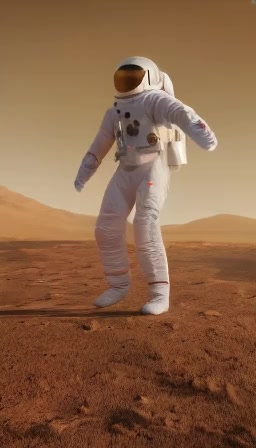}
\videoframe{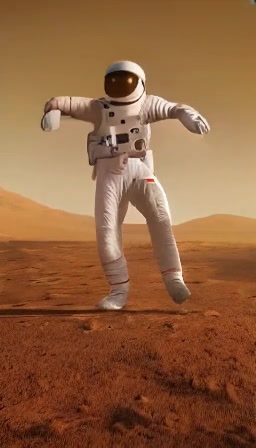}
\videoframe{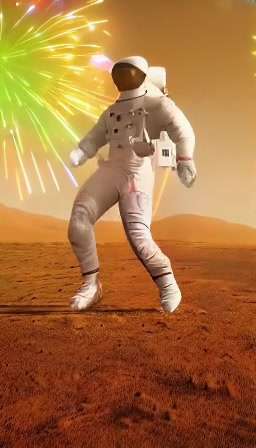}
\videoframe{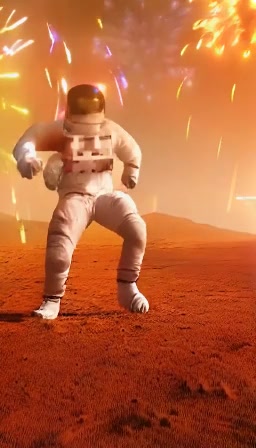}

\end{subfigure}

\captionof{figure}{\textbf{10-Second long video generation example.} By predicting 1-second video segments from an initial 1-second clip, \modelname{} can iteratively generate videos of extended lengths.}
\label{fig:long_video}
\vspace{-4mm}
\end{figure}

\begin{figure}[t]
\addtocounter{figure}{-1}


\begin{subfigure}{1.0\linewidth}
\centering
\videoframe{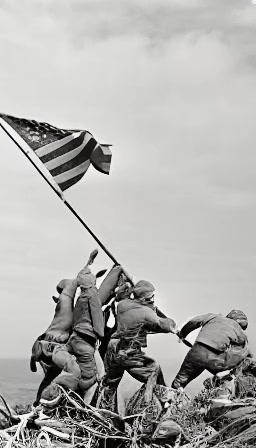}
\videoframe{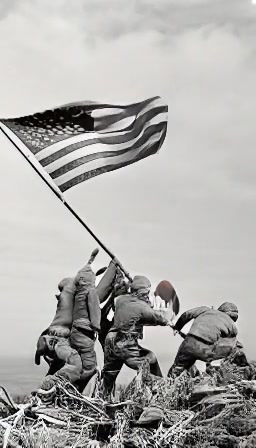}
\videoframe{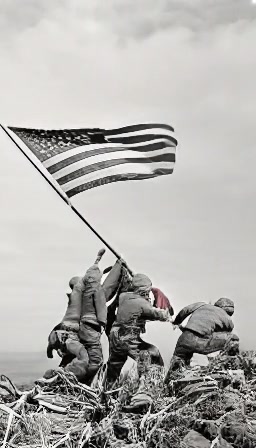}
\videoframe{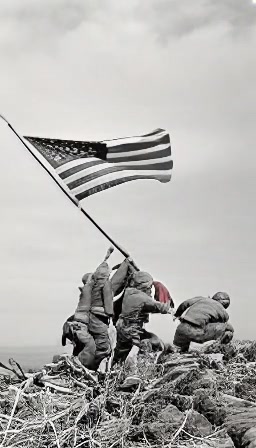}
{\textbf{Animated from historical photo}
\ \\
\ \\}
\end{subfigure}

\begin{subfigure}{1.0\linewidth}
\centering
\videoframe{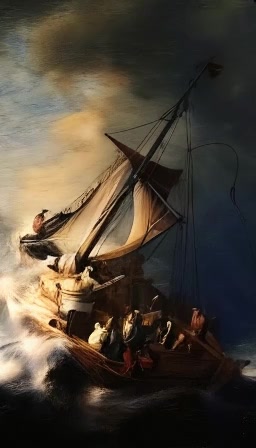}
\videoframe{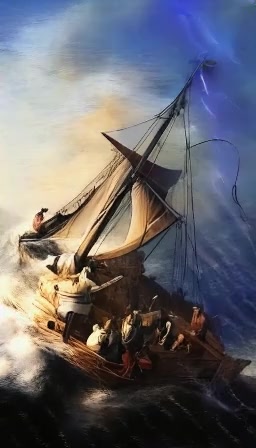}
\videoframe{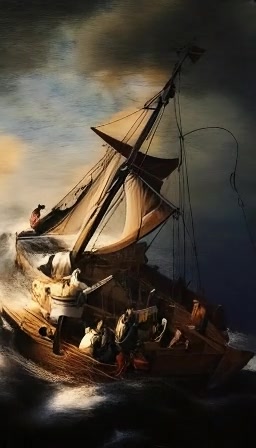}
\videoframe{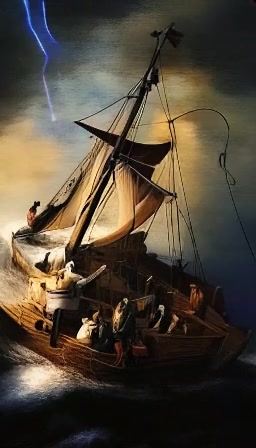}
{\textbf{Animated from painting}
\ \\}
\end{subfigure}
\captionof{figure}{\textbf{Examples of videos animated from still images plus text prompts tailored to each initial image.}}
\label{fig:animated_images}
\end{figure}

A benefit of an decoder-based language model is that it pairs well with autoregressively extending generation in time. 
We present two different variants: Generating longer videos and converting images to videos. 
Encoding the first frame independently allows us to convert any image into the initial frame of a video without padding. Subsequent frames are generated by predicting remaining tokens, transforming the image into a video as shown in \Cref{fig:animated_images}\footnote{For image-to-video examples we source images from Wikimedia Commons: https://commons.wikimedia.org/wiki/Main\_Page}.

This results in the capability to generate videos longer than 10 seconds or to allow users to iteratively extend video clips based on previously generated video, and produces temporally consistent videos without significant distortion.
Such capabilities are rarely observed in contemporary diffusion models.

\vspace{-4mm}
\paragraph{Zero-shot video editing and task chaining.} \label{sec:capabilities}
\begin{figure}[tp]
\addtocounter{figure}{-1}
\centering
\begin{subfigure}{1.0\linewidth}
\centering
\videoframe{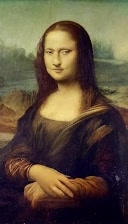}
\videoframe{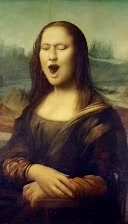}
\videoframe{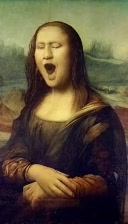}
\videoframe{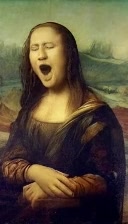}
{\textbf{Animated from still image}
\ \\
\ \\}
\end{subfigure}

\begin{subfigure}{1.0\linewidth}
\centering
\videoframe{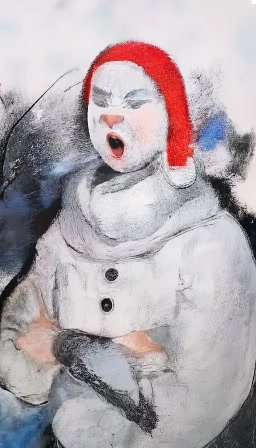}
\videoframe{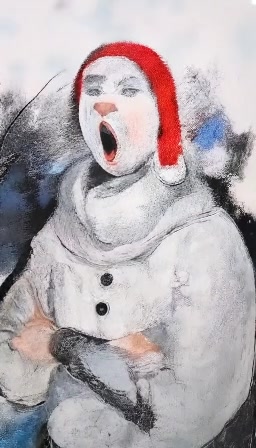}
\videoframe{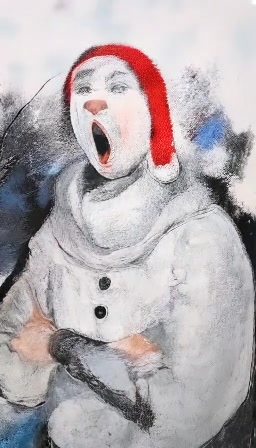}
\videoframe{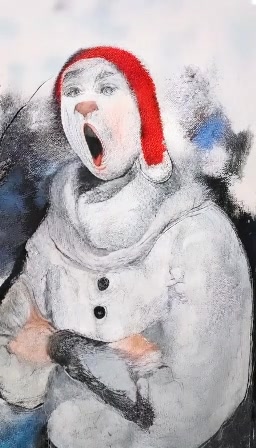}
{\textbf{Stylized video} \\
\textbf{Prompt:} An oil painting of a snowman with a red hat opening their mouth to yawn
\ \\}
\end{subfigure}
\captionof{figure}{\textbf{Example of zero-shot video editing via task chaining} (text conditioned image-to-video and stylization) 
}
\label{fig:animate_stylize}
\vspace{-2mm}
\end{figure}

With the multi-task pretraining, \modelname{} exhibits task generalization that can be chained together to perform novel tasks.
We show the model can apply image-to-video animation followed by video-to-video stylization in \Cref{fig:animate_stylize}.
In the Appendix, \Cref{fig:outpaint_stylize} shows another example applying video-to-video outpainting, followed by editing them with additional video-to-video effects.
%
At each stage, the quality of the output is sufficient to remain in-distribution (\ie teacher forcing) for the next stage without noticeable artifacts.
These capabilities can be attributed to our multimodal task design within an LLM transformer framework that allows for modeling multimodal content using a single transformer architecture over a unified vocabulary. 

\vspace{-4mm}
\paragraph{Zero-shot video stylization.}
Stylization results are presented in \Cref{sec:stylization} where the structure and text are used as prefixes to guide the language model. Unlike other stylization methods that employ adapter modules such as cross-attention networks~\cite{zhang2023adding} or latent blending~\cite{meng2021sdedit}, our approach stylizes videos within an LLM as one of several generative tasks.

\vspace{-4mm}
\paragraph{3D structure, camera motion, visual styles.}
Even though we do not add specific training data or losses to encourage 3D consistency, our model
can rotate around objects and predict reasonable visualizations of the backside of objects. Additionally, with only a small proportion of input videos and texts describing camera motion, our model can use short text prompts to apply a range of camera motions to image-to-video and text-to-video generations (see \Cref{fig:camera_movement}).

\subsection{Limitations}

Despite VideoPoet demonstrating highly competitive performance of LLMs relative to state-of-the-art models, certain limitations are still observed.
For example, the RGB frame reconstruction from compressed and quantized tokens place an upper bound on the generative model's visual fidelity. Second, the per-frame aesthetic biases in static scenes does not match the best baseline.
This difference is largely due to a choice of training data, where we focus our training on more natural aesthetics 
and excluded some sources containing copyrighted images, such as LAION~\cite{schuhmann2022laion}, which is commonly used in other work.
Finally, small objects and fine-grained details, especially when coupled with significant motions, remains difficult within the token-based modeling.



\vspace{-2mm}
\section{Conclusion}
\vspace{-1mm}

\modelname{} demonstrates the potential of a large language model that is trained on discrete visual, text and audio tokens, in generating videos of compelling state-of-the-art quality.
A particular strength of our model lies in its ability to
generate high-fidelity, large, and complex motions.
Our large language model formulation benefits from training
across a variety of multimodal tasks with a unified architecture and vocabulary.
Consequently, the pretrained model is adept at multi-task video creation, and serves as a foundation for a diverse variety of video generation related capabilities, including multiple forms of editing.


\section*{Acknowledgements}
We give special thanks to Alex Siegman, Victor Gomes, and Brendan Jou for managing computing resources.
We also give thanks to Aren Jansen, Marco Tagliasacchi, Neil Zeghidour, John Hershey for audio tokenization and processing, Angad Singh for storyboarding in ``Rookie the Raccoon'', Cordelia Schmid for research discussions, David Salesin, Tomas Izo, and Rahul Sukthankar for their support, and Jay Yagnik for the initial concept.

\section*{Impact Statement}
This paper introduces research aimed at advancing the field of video generation and introduces a new tool to enhance human creativity. In considering the societal impact of our video generation model, VideoPoet, several concerns emerge, including the potential for misuse and ethical considerations. Misuse may entail the creation of deceptive or harmful content, such as misinformation or deepfakes. Ethical considerations encompass ensuring that generated content adheres to ethical standards, avoids perpetuating harmful stereotypes, and respects cultural sensitivities. To mitigate these risks, we adhere to evolving strategies developed within our community. This includes using digital watermarking in generated videos to enable traceability and accountability, and maintain transparency in our model design to foster trust.

{
    \small
    \bibliographystyle{icml2024}
    \bibliography{main}
}

\newpage
\appendix
\onecolumn

\section{Appendix} \label{sec:appendix}

\subsection{Responsible AI and Fairness Analysis}

We evaluate whether the generated outputs of our model are fair regarding protected attributes such as (1) Perceived Age (2) Perceived Gender Expression (3) Perceived Skin Tone.
We construct 306 prompts with template --- ``a \{profession or people descriptor\} looking \{adverb\} at the camera” with ``profession'' being crawled from the US Bureau of Labor and Statistics and ``people descriptors'' including emotion state, socioeconomic class, \etc.
The ``adverb'' is used to generate semantically unchanged prompt templates such as ``straightly'' or ``directly''. 
We generate 8 videos for each prompt and for each generated video we infer an approximation of the expressed attribute regarding the 3 protected attributes.
Across 10 prompts that have the same semantic meaning but different ``adverbs'', we observe our outputs generally introduced a stronger distribution shift toward ``Young Adults'' (age 18-35), ``Male'' and ``Light Skin Tone''.
However, we observe changing the ``adverb'' in the prompt template can significantly alter the output distributions. Therefore, our model can be prompted to produce outputs with non-uniform distributions across these groups, but also possess the ability of being prompted to enhance uniformity, though prompts are semantically unchanged.
While research has been conducted in the image generation and recognition domain~\cite{zhang2023auditing, schumann2021step, schumann2023consensus, Chiu_2023_ICCV}, 
this finding highlights the importance of continued research to develop strategies to mitigate issues and improve fairness for video generation.

\subsection{Model Scale and Performance}
To analyze model performance versus model scale, we use a subset of the training set without text-paired data and a slightly different task prompt design. 
We evaluate the video generation quality using FVD~\cite{unterthiner2018towards} and audio generation quality using the Fréchet Audio Distance (FAD), which uses the VGGish model as the embedding function~\cite{hershey2017cnn}. 
Both FVD and FAD metrics are calculated using a held-out subset of 25 thousand videos.

\Cref{fig:model_scale} shows that as the model size grows and the amount of training data increases, performance improves across visual and audiovisual tasks.
After obtaining the above results, we retrain our 1B and 8B models using the task design and text-paired training data discussed in \Cref{sec:method}.
\Cref{sec:appendix-sxs} shows a qualitative comparison of the 1B and 8B pretrained models. Increasing the model size improved temporal consistency, prompt fidelity, and motion dynamics while adding capabilities for limited text rendering, spatial understanding, and counting.

\begin{figure*}[h]
\centering
\begin{subfigure}{0.48\textwidth}  
\includegraphics[width=\linewidth]{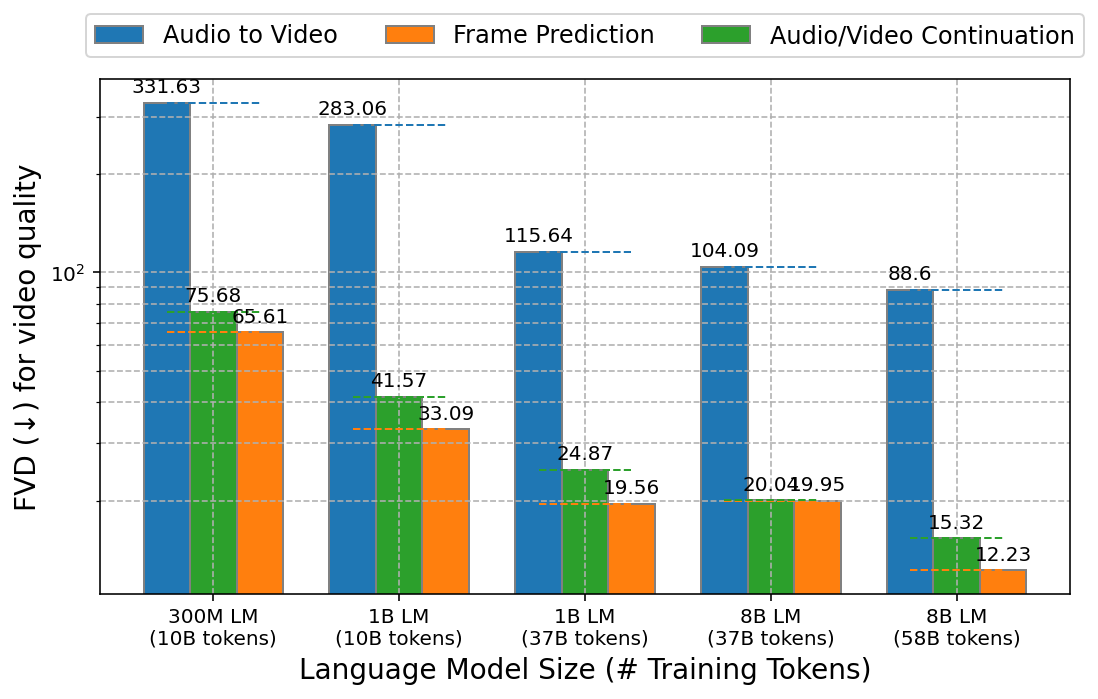}
\subcaption{Video generation quality in FVD ($\downarrow$).}
\end{subfigure}
\begin{subfigure}{0.48\textwidth}  
\includegraphics[width=\linewidth]{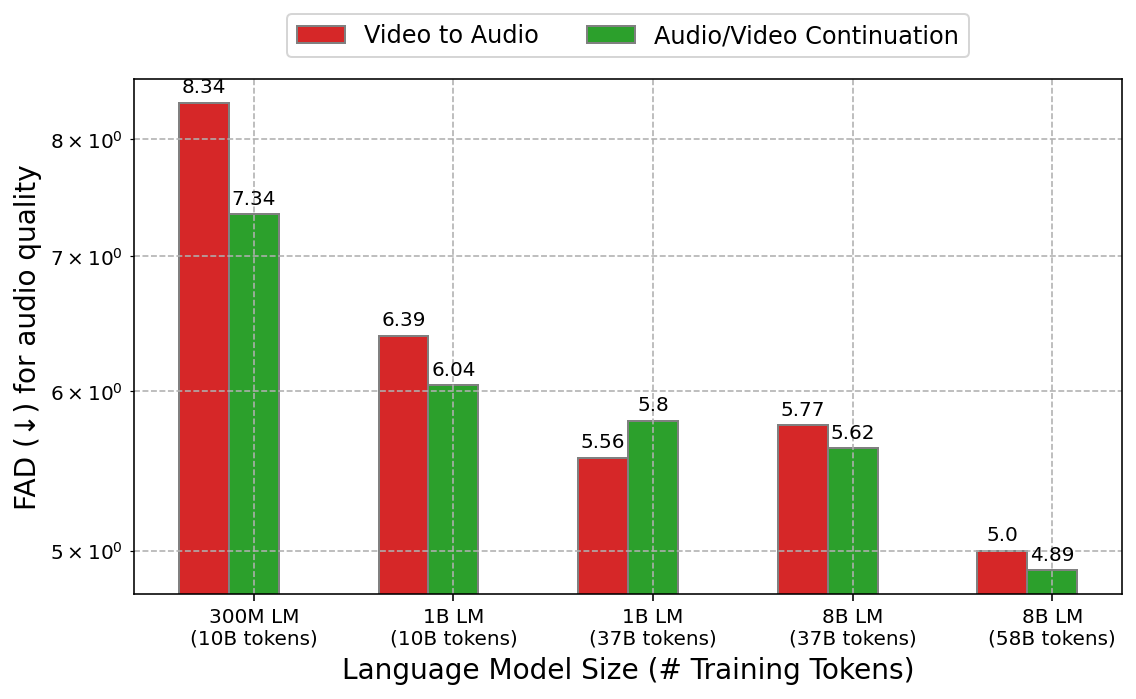}
\subcaption{\label{fig:model_scale_b}Audio generation quality in FAD  ($\downarrow$).}
\end{subfigure}
\caption{\textbf{Effects of model and data scale on video and audio generation quality}. The performance, depicted on a log-log scale, improves significantly when we scale up the model and training data. 
Language models with 300 million, 1 billion, and 8 billion parameters are trained on datasets comprising 10, 37, and 58 billion visual and audio tokens, respectively.
} 
\label{fig:model_scale}
\end{figure*}

\subsubsection{Qualitative Comparison of 1B and 8B models} \label{sec:appendix-sxs}

\begin{figure*}[tp]
\addtocounter{figure}{-1}

\begin{subfigure}{1.0\linewidth}
\centering
\includegraphics[width=0.11\textwidth]{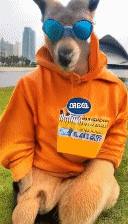}
\includegraphics[width=0.11\textwidth]{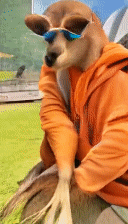}
\includegraphics[width=0.11\textwidth]{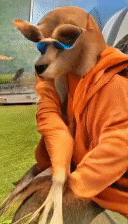}
\includegraphics[width=0.11\textwidth]{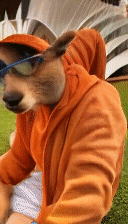}
\hspace{0.03\textwidth}
\includegraphics[width=0.11\textwidth]{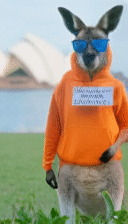}
\includegraphics[width=0.11\textwidth]{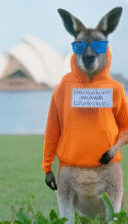}
\includegraphics[width=0.11\textwidth]{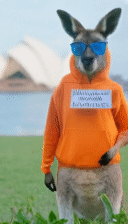}
\includegraphics[width=0.11\textwidth]{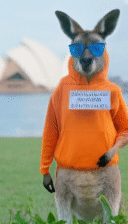}
~
{\textbf{prompt:} A portrait photo of a kangaroo wearing an orange hoodie and blue sunglasses standing on the grass
in front of the Sydney Opera House holding a sign on the chest that says Welcome Friends!
\ \\
\ \\}

\end{subfigure}

~

\begin{subfigure}{1.0\linewidth}
\centering
\includegraphics[width=0.11\textwidth]{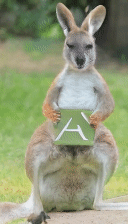}
\includegraphics[width=0.11\textwidth]{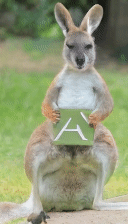}
\includegraphics[width=0.11\textwidth]{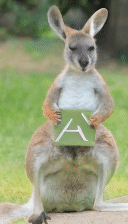}
\includegraphics[width=0.11\textwidth]{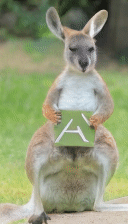}
\hspace{0.03\textwidth}
\includegraphics[width=0.11\textwidth]{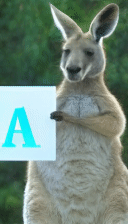}
\includegraphics[width=0.11\textwidth]{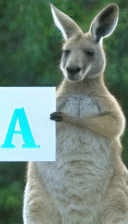}
\includegraphics[width=0.11\textwidth]{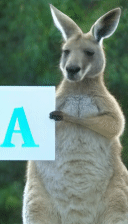}
\includegraphics[width=0.11\textwidth]{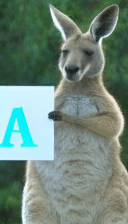}
{\textbf{prompt:} A kangaroo holding a sign with the letter A on it
\ \\
\ \\}

\end{subfigure}

~

\begin{subfigure}{1.0\linewidth}
\centering
\includegraphics[width=0.11\textwidth]{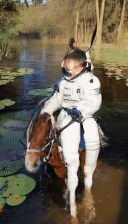}
\includegraphics[width=0.11\textwidth]{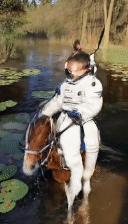}
\includegraphics[width=0.11\textwidth]{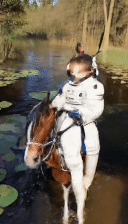}
\includegraphics[width=0.11\textwidth]{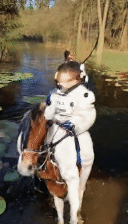}
\hspace{0.03\textwidth}
\includegraphics[width=0.11\textwidth]{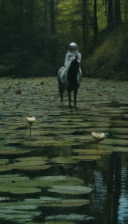}
\includegraphics[width=0.11\textwidth]{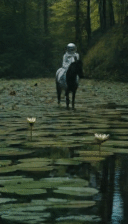}
\includegraphics[width=0.11\textwidth]{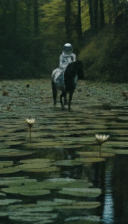}
\includegraphics[width=0.11\textwidth]{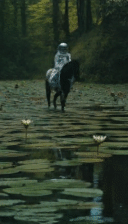}
{\textbf{prompt:} A photo of an astronaut riding a horse in the forest. There is a river in front of them with water lilies
\ \\
\ \\}

\end{subfigure}

~

\begin{subfigure}{1.0\linewidth}
\centering
\includegraphics[width=0.11\textwidth]{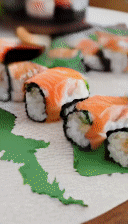}
\includegraphics[width=0.11\textwidth]{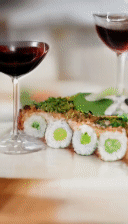}
\includegraphics[width=0.11\textwidth]{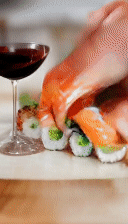}
\includegraphics[width=0.11\textwidth]{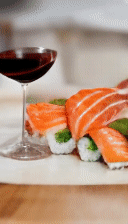}
\hspace{0.03\textwidth}
\includegraphics[width=0.11\textwidth]{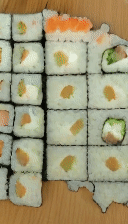}
\includegraphics[width=0.11\textwidth]{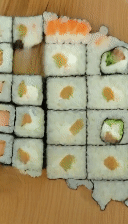}
\includegraphics[width=0.11\textwidth]{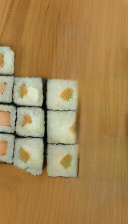}
\includegraphics[width=0.11\textwidth]{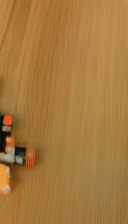}
{\textbf{prompt:} A zoomed out map of the United States made out of sushi. It is on a table next to a glass of red wine. Pieces of sushi disappear one by one
\ \\
\ \\}

\end{subfigure}

~

\begin{subfigure}{1.0\linewidth}
\centering
\includegraphics[width=0.11\textwidth]{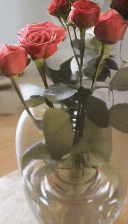}
\includegraphics[width=0.11\textwidth]{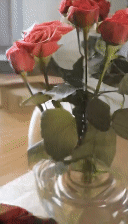}
\includegraphics[width=0.11\textwidth]{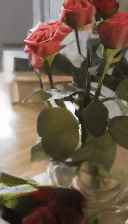}
\includegraphics[width=0.11\textwidth]{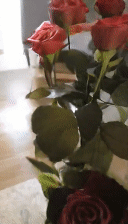}
\hspace{0.03\textwidth}
\includegraphics[width=0.11\textwidth]{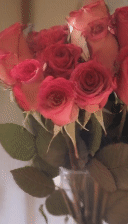}
\includegraphics[width=0.11\textwidth]{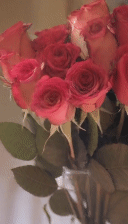}
\includegraphics[width=0.11\textwidth]{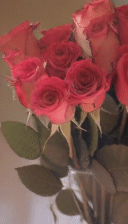}
\includegraphics[width=0.11\textwidth]{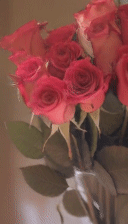}
{\textbf{prompt:} Rotating around a vase holding a dozen roses
\ \\
\ \\}

\end{subfigure}

\captionof{figure}{\textbf{A comparison of a 1B (left) and 8B (right) parameter models} on the same prompt and settings.}

\label{fig:sxs_1b_8b}
\end{figure*}

In Figure~\ref{fig:sxs_1b_8b}, we show outputs of 1B and 8B parameter models on the same prompts.
Four frames from the best video output of each model in a batch of four text-to-video samples were selected to represent the model.
In the first row, the 1B model is unstable with large changes to the subject over time and misses elements from the complex prompt.
This prompt was originally used for scaling comparisons in~\cite{yu2022scaling}, and compared to a dedicated image-only model, our model does not preserve text as well given the training data used. In the second row, we use a simpler text task and show that the 8B model can represent a single letter clearly, but the 1B model still produces artifacts.
In the third row, we show that the 8B model learns spatial positioning such as the river being in front of the astronaut and horse. In the fourth row, we show that the 8B parameter model learned a stop motion style to have items disappear ``one by one" and can follow a complicated layout from a long prompt.
In contrast, the 1B model includes all of the nouns, but is unstable over time and does not follow the layout indicated in the prompt. In the bottom row, we show that the 8B model understands counts of objects in that it displays a full bouquet (though 12 roses are not explicitly in frame) and smooth consistent motion as opposed to the 5 roses and distorting objects produced by the 1B model. 
Overall, scaling the model improved temporal consistency, prompt fidelity, and motion dynamics while adding capabilities for limited text rendering, spatial understanding, and counting.


\subsection{Additional Generated Examples} 
We include most generated videos in the supplementary materials for an enhanced visualization of motion and visual quality, in addition to \Cref{fig:outpaint_stylize} and \Cref{fig:camera_movement}.


\begin{figure}[tp]
\centering
\begin{subfigure}[b]{0.4\linewidth}
\centering
\videoframe{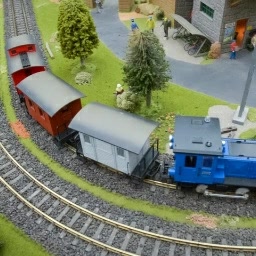}
\videoframe{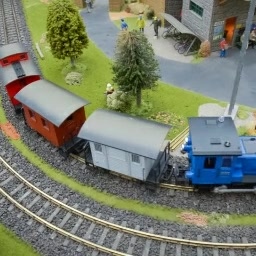}
\videoframe{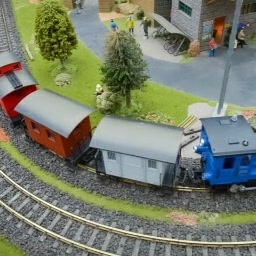}
\videoframe{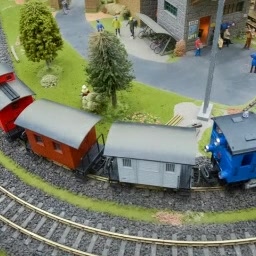}
{\textbf{Original Video}
\ \\
\ \\}
\end{subfigure}

\begin{subfigure}[b]{0.4\linewidth}
\centering
\videoframe{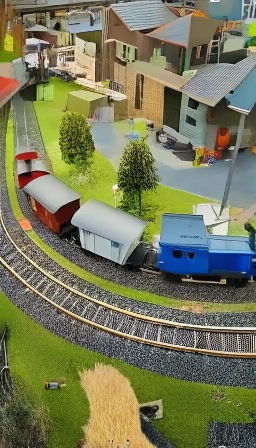}
\videoframe{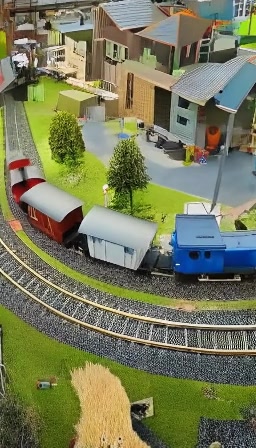}
\videoframe{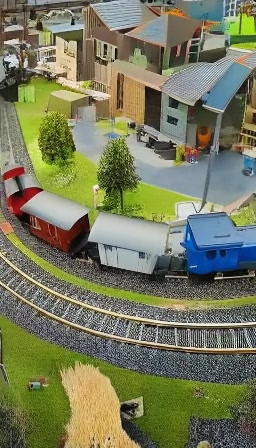}
\videoframe{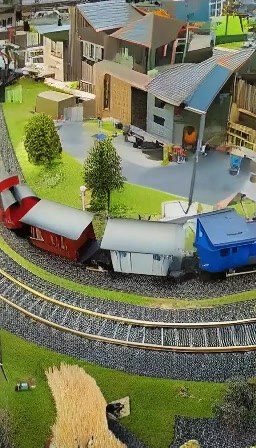}
{\textbf{Outpainted Video}
\ \\
\ \\}
\end{subfigure}

\begin{subfigure}[b]{0.4\linewidth}
\centering
\videoframe{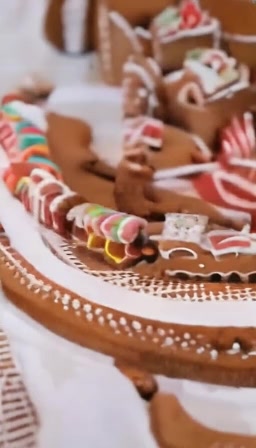}
\videoframe{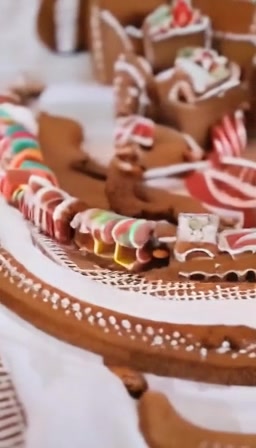}
\videoframe{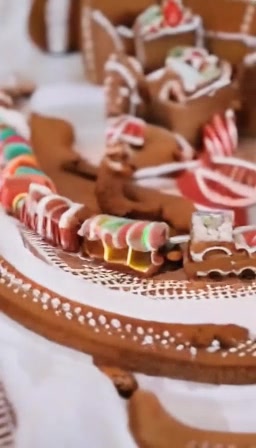}
\videoframe{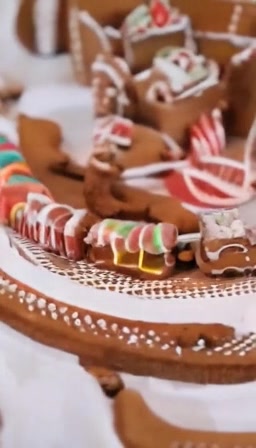}
{\textbf{Stylized Video} \\
\textbf{Prompt:} A gingerbread and candy train on a track
\ \\}
\end{subfigure}

\caption{\textbf{Example of zero-shot video editing via task chaining (outpainting and stylization)} -- the original video is first outpainted and then stylized via a text prompt.}
\label{fig:outpaint_stylize}
\end{figure}

\begin{figure}[tp]
\addtocounter{figure}{-1}
\centering
\begin{subfigure}{0.4\linewidth}

\centering
\videoframe{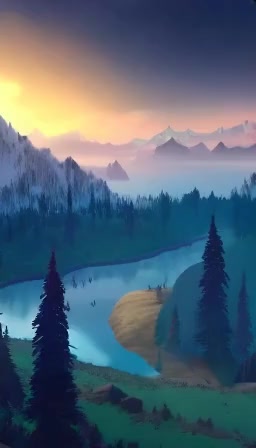}
\videoframe{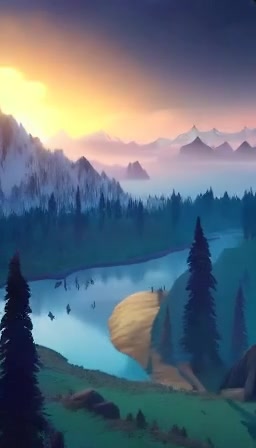}
\videoframe{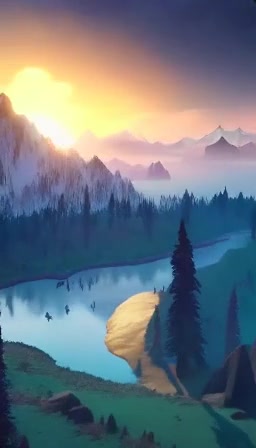}
\videoframe{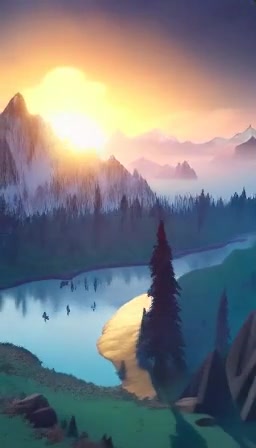}
{\textbf{ Camera Motion}: Arc shot
\ \\
\ \\}
\end{subfigure}

\begin{subfigure}{0.4\linewidth}
\centering
\videoframe{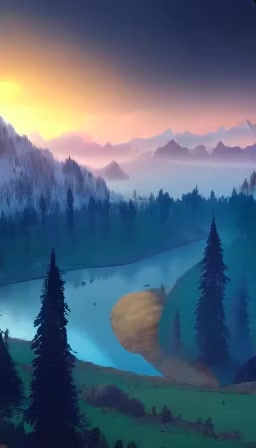}
\videoframe{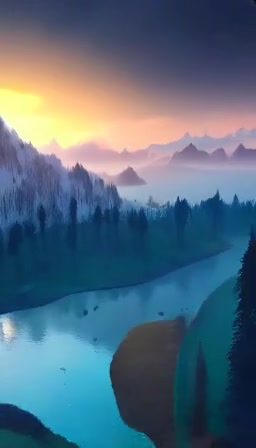}
\videoframe{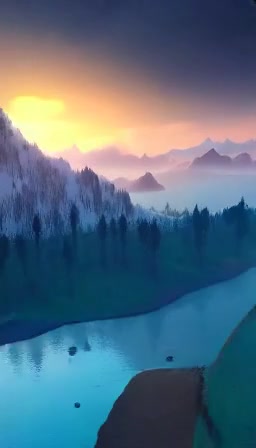}
\videoframe{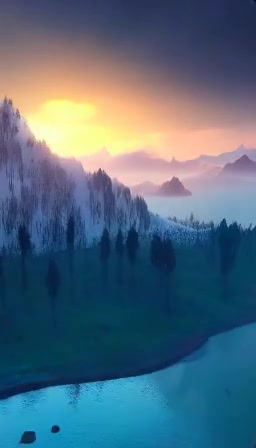}
{\textbf{ Camera Motion}: FPV drone shot
\ \\}
\end{subfigure}

\captionof{figure}{\textbf{Examples of directed camera movement from the same initial frame.}}
\label{fig:camera_movement}
\end{figure}


\subsection{Video Stylization}
\label{sec:stylization}
To perform video stylization, we follow an approach inspired by~\cite{zhang2023adding,chen2023control,esser2023structure} to predict videos from the combination of text, optical flow, and depth signals. On a subset of steps, we also condition on the first video frame. As described in~\cite{esser2023structure}, the text will generally define the ``content" or appearance of the output and the optical flow and depth control the ``structure." In contrast to the diffusion-based approaches that usually use external cross-attention networks~\cite{zhang2023adding} or latent blending~\cite{meng2021sdedit} for stylization, our approach is more closely related to machine translation using large language models in that we only need to provide the structure and text as a prefix to a language model. 

To perform the task, we estimate optical flow from RAFT~\cite{sun2022disentangling} and produce monocular depth maps from MIDAS~\cite{ranftl2020midas}, and then normalize and concatenate on the channel dimension. This conveniently produces the same number of channels as the RGB ground truth and so can be tokenized in the same fashion as RGB videos with the MAGVIT-v2 tokenizer without retraining the tokenizer. The task of stylization is to reconstruct the ground truth video from the given optical flow, depth, and text information.
During inference, we apply optical flow and depth estimation on an input video but then vary the text prompt to generate a new style, \eg ``cartoon''.

\begin{table}[h]
\centering
\caption{\textbf{Comparison on video stylization.} \modelname{} outperforms Control-A-Video by a large margin. 
\label{table:stylization}
}
\begin{tabular}{lc}
\toprule
Model & CLIPSIM \\
\midrule
Control-A-Video~\cite{chen2023control}[depth] & 0.3246 \\
\textbf{\modelname{} (Ours)} & 0.3417 \\
\bottomrule
\end{tabular}
\end{table}

\begin{figure}
\centering
\begin{subfigure}{0.6\textwidth}  
\includegraphics[width=0.99\linewidth]{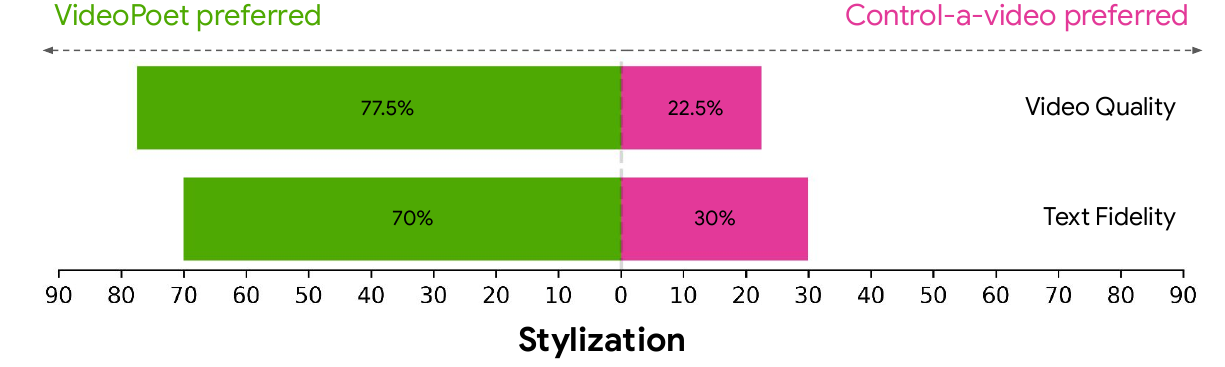}
\end{subfigure}
\caption{\textbf{Human side-by-side evaluations comparing \modelname{} with the video stylization model Control-a-video~\cite{chen2023control}.} Raters prefer \modelname{} on both text fidelity and video quality.
Green and pink bars represent the proportion of trials where \modelname{} was preferred over an alternative, or preferred less than an alternative, respectively.
}
\label{fig:human_t2v_evals_stylization}
\end{figure}

To evaluate stylization capabilities, we choose 20 videos from the public DAVIS 2016\footnote{DAVIS license: \url{https://creativecommons.org/licenses/by-nc/4.0/deed.en}}~\cite{Perazzi2016davis} dataset and provide 2 style prompts for each video. For more details, please refer to \Cref{sec:appendix-stylization-davis}. Following~\cite{esser2023structure},
we evaluated the CLIP-embedding consistency between each frame and the text prompt
to determine if the stylization results matches the text. As shown in \Cref{table:stylization}, \modelname{} outperforms Control-A-Video conditioned on depth by a large margin. We also conduct human evaluations as discussed above comparing with Control-A-Video~\cite{chen2023control}.
Human raters consistently 
prefer our text fidelity and video quality as shown in \Cref{fig:human_t2v_evals_stylization}.

\subsection{Additional Implementation and Evaluation Details}


\subsubsection{Additional Implementation Details}
The unified vocabulary is constructed as follows: the initial 256 codes are reserved for special tokens and task prompts. \Cref{tab:special_tokens} lists some examples of special tokens. Subsequently, the next 262,144 codes are allocated for image and video tokenization. This is followed by 4,096 audio codes. We also include a small text vocabulary of English words. Overall, this produces a total vocabulary size of approximately 300,000.


\begin{table}
\centering
\begin{tabular}{cl}
\toprule
Special Token                          & \multicolumn{1}{c}{Usage} \\
\midrule
\texttt{<bos>} & Beginning of sequence     \\
\texttt{<task>} & Task to perform for this sequence \\
\texttt{<bot\_i>}& Beginning of the text input. \\
\texttt{<eot\_i>}& End of the text input. \\
\texttt{<bov\_i>}& Beginning of the visual input. \\
\texttt{<eov\_i>}& End of the video input. \\
\texttt{<boa\_i>}& Beginning of the audio input. \\
\texttt{<eoa\_i>}& End of the audio input. \\
\texttt{<source>}& The source of the video to generate. \\
\texttt{<res>} & Output resolution for the video. \\
\texttt{<bov\_o>}& Beginning of the video output. \\
\texttt{<eov\_o>}& End of the video output. \\
\texttt{<boa\_o>}& Beginning of the audio output. \\
\texttt{<eoa\_o>}& End of the audio output. \\
\texttt{<eos>}& End of the entire sequence. \\
\bottomrule
\end{tabular}
\caption{\textbf{List of representative special tokens used in training and inference}. }
\label{tab:special_tokens}
\end{table}

Since the first frame is tokenized separately, MAGVIT-v2 allows images to be represented in the same vocabulary as video.
In addition to being more compact, images provide many learnable characteristics that are not typically represented in videos, such as strong visual styles (\eg, art paintings), objects which are infrequently seen in video, rich captions, and significantly more text-image paired training data.
When training on images, we resize the images to 128$\times$128 which are then tokenized to a latent shape of $(1, 16, 16)$, or 256 tokens.
We scale the MAGVIT-v2 model's size and train it on the datasets discussed in \Cref{sec:exp_data}. The training follows two steps: image training, inflation~\cite{yu2023language} and video training. Due to images requiring fewer tokens, we can include roughly 5$\times$ more images per batch than videos, \ie 256 image tokens vs. 1280 video tokens. We use up to a maximum of 64 text tokens for all of our experiments. For the \texttt{<res>} token, the resolution is only specified for $128\times224$ output, $128\times128$ resolution is assumed otherwise.

The video-to-video tasks use the COMMIT encoding~\cite{yu2023magvit} to obtain the tokens for the tasks such as inpainting and outpainting.
Text is encoded as T5 XL embeddings~\cite{raffel2020exploring} and are inserted into reserved sequence positions right after the \texttt{<bot\_i>} token as shown in \Cref{fig:vffm_schematic}.
 

\subsubsection{Super-resolution implementation details} \label{sec:appendix-superres-details}
We use a 1B model for the first $2\times$ spatial super-resolution stage and a 500M model for the second $2\times$ stage.
The first super-resolution stage models videos of $17\times448\times256$ pixels with a  token sequence of shape $(5, 56, 32)$. The second stage models videos of $17\times896\times512$ pixels with a token sequence of shape $(5, 112, 64)$. The token sequences are obtained with the same MAGVIT-v2~\citep{yu2023language} tokenizer used for the base language model.
The custom super-resolution transformer has local self-attention windows for \emph{vertical}, \emph{horizontal} and \emph{temporal} layers of shape $(1, 56, 4), (1, 8, 32), (5, 8, 8)$ in the first stage and  $(1, 112, 2), (1, 4, 64), (5, 8, 8)$ in the second stage, respectively (\Cref{fig:sr_diagram}). The cross-attention layers attend to local windows in the low-resolution sequence isomorphic to self-attention windows but with half the spatial size.

We train the super-resolution stages on a dataset of 64M high-quality text-video  pairs using the masked modeling objective of MAGVIT~\citep{yu2023magvit}, with token factorization into $k=2$ groups~\cite{yu2023language}. During inference, we use the sampling algorithm of MAGVIT-v2~\citep{yu2023language} with 24 sampling steps for each stage and classifier-free guidance scale~\citep{ho2022classifier,brooks2023instructpix2pix} of $4.0/8.0$ for the text condition and $1.0/2.0$ for the low-resolution condition, in the first/second stage.

\subsubsection{Additional Evaluation details}

We measure CLIP similarity scores~\cite{wu2021godiva} following an implementation given by \citet{villegas2022phenaki}, measure FVD~\cite{unterthiner2018towards} following \citet{yu2023magvit} on UCF101 dataset and following \citet{zhang2023show} on MSR-VTT, and measure Inception Score (IS)~\cite{saito2020train}. When the evaluation protocol is on 16 frames, we discard the generated last frame to make a 16-frame video.

\subsubsection{Additional Human Evaluation details}
\label{sec:appendix-human-evaluation}

\begin{figure*}[t]
\centering
\includegraphics[width=0.75\textwidth]{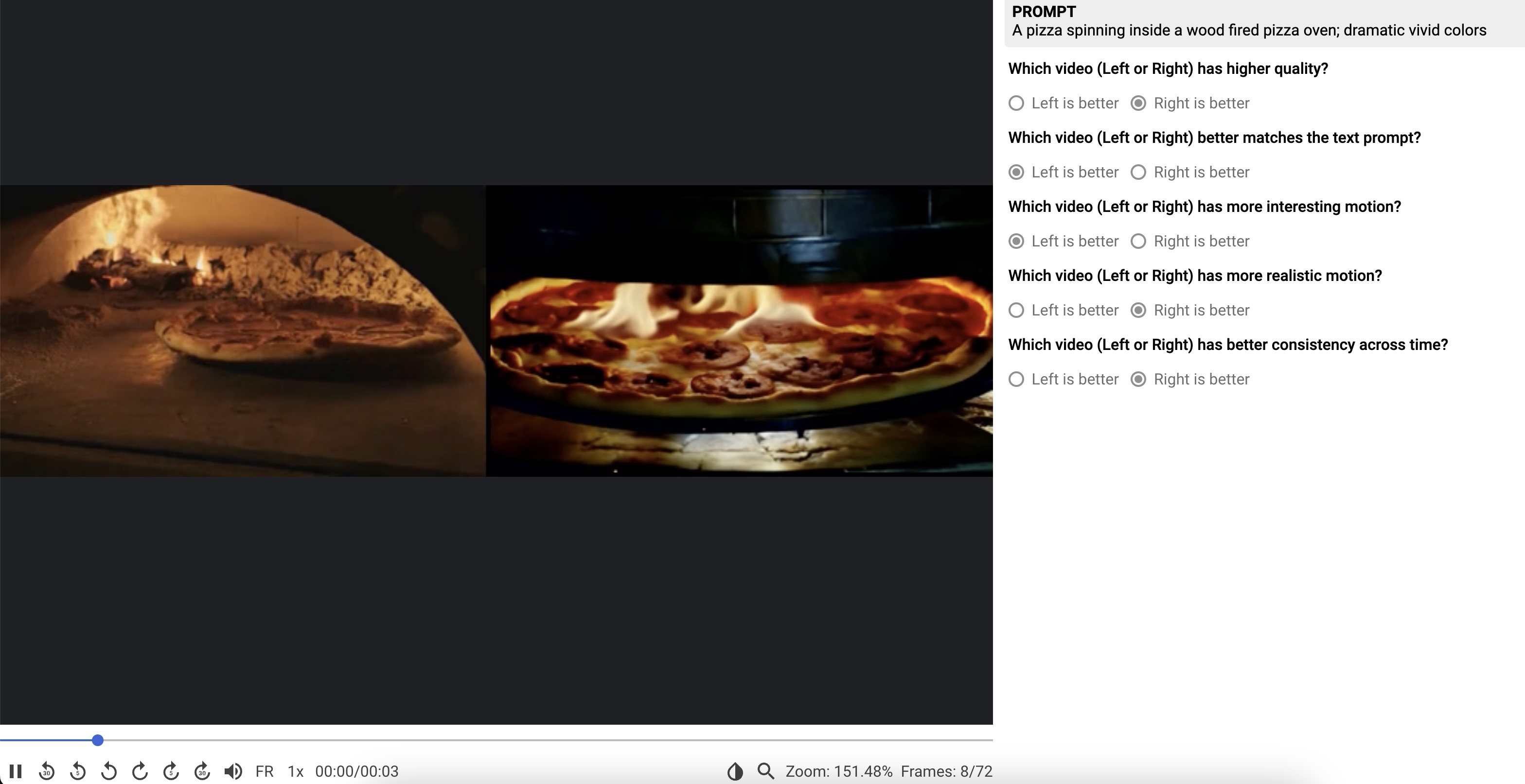}
\caption{Example
screenshot of the user interface for human
side-by-side comparisons.}
\label{fig:human_eval_screenshot}
\end{figure*}

Figure~\ref{fig:human_eval_screenshot} shows an example screenshot
of our side-by-side UI for comparing models.  We used a team of 7
human raters to complete all ratings.  To achieve best results, we
call VideoPoet using the negative prompt \emph{``a
still shot of an ugly cartoon, slideshow of an empty scene,
low resolution, distorted and disfigured''}
and rewrite the given prompt by appending the string
\emph{highly detailed, cinematic, arc shot, high contrast, soft lighting, 8k}.

\subsubsection{Zero-shot Text-to-Video Evaluation Settings}
\label{sec:appendix-evaluation}
We report the details of our zero-shot text-to-video settings here.
We note that some details are missing in previous papers and different papers use different settings. Hence, we provide all the details and hope this evaluation setting can serve as a standard text-to-video generation benchmark.
Our results are reported on the 8B model and we adopt classifier-free guidance~\cite{ho2022classifier}.

\vspace{-2mm}
All metrics are evaluated on generated videos containing 16 frames with a resolution of 256 x 256.
We first generate videos of 128 x 128 resolution and then resize to 256 x 256 via bicubic upsampling.

\emph{Zero-shot MSR-VTT}.
For CLIP score, we used all 59,794 captions from the MSR-VTT test set. We use CLIP ViT-B/16 model following Phenaki~\cite{villegas2022phenaki}.
We note that some papers use other CLIP models, \eg, VideoLDM~\cite{blattmann2023align} uses ViT-B/32.
Our CLIP score evaluated on the ViT-B/32 backbone for MSR-VTT is 30.01.
For the FVD metric, to evaluate on a wide range of captions as well as to be comparable with previous papers that evaluate on 2,048 videos, we evaluate on the first 40,960 captions in the MSR-VTT test set. More specifically, we report the FVD metrics on 2048 videos with 20 repeats.
The FVD real features are extracted from 2,048 videos sampled from the MSR-VTT test set.
We sample the central 16 frames of each real video, without any temporal downsampling, \ie, we use the original fps in the MSR-VTT dataset (30 fps as reported in \citet{xu2016msr}). The FVD is evaluated with an I3D model trained on Kinetics-400.

\emph{Zero-shot UCF-101}.
Following VDM~\cite{ho2022video}, we sample 10,000 videos from the UCF-101 test set and use their categories as the text prompts to generate 10,000 videos.
We use the class text prompts provided in PYoCo~\cite{ge2023preserve} to represent the 101 categories.
To compute the FVD real features, we sample 10K videos from the training set, following TGAN2~\cite{saito2020train}.
We sample the central 16 frames for each real video , without any temporal downsampling, \ie, we use the original fps in the UCF-101 dataset (25 fps as reported in \cite{soomro2012ucf101}).
The FVD metric is evaluated with an I3D model trained on Kinetics-400 and the IS metric is evaluated with a C3D model trained on UCF-101.

\subsubsection{Self-Supervised Tasks Evaluation Settings}
\label{sec:ssl_task}
Self-supervised learning tasks include frame prediction on K600 with 5 frames as condition, as well as inpainting and outpainting on SSv2.
FVD~\cite{unterthiner2018towards} is used as the primary metric, calculated with 16 frames at 128$\times$128 resolution.
We follow MAGVIT~\cite{yu2023magvit} in evaluating these tasks against the respective real distribution, using 50000$\times$4 samples for K600 and 50000 samples for SSv2.

\subsubsection{Stylization Evaluation on DAVIS} \label{sec:appendix-stylization-davis}

To evaluate the CLIP similarity score and human preference on video stylization, we use the following set of videos and prompts. We select 20 videos from DAVIS 2016~\cite{Perazzi2016davis}, and for each video we take 16 frames starting from the initial frame specified below and evaluate stylization on the two text prompts specified below. To be easily reproducible, we use a central square crop at the height of the video and evaluate the output videos at 256x256 resolution. We use CLIP-B/16 for the similarity score. Several prompts below are used in or inspired by previous work~\cite{esser2023structure,chen2023control,liew2023magicedit}.

\begin{table*}[tp]
\centering
\begin{tabular}{lcp{10cm}}
\toprule
video name & starting frame & first text prompt \\
\midrule
elephant & 10 & oil painting of an elephant walking away \\
elephant & 10 & cartoon animation of an elephant walking through dirt surrounded by boulders \\
car-turn & 40 & car on a snowcovered road in the countryside \\
car-turn & 40 & 8-bit pixelated car driving down the road \\
dog-agility & 0 & a dog in the style of a comic book \\
dog-agility & 0 & a dog running through a field of poles in the style of cyberpunk \\
bmx-bumps & 10 & riding a bicycle on a rainbow track in space with stars and planets in the background \\
bmx-bumps & 10 & riding a bicycle on a dirt track in the style of a graphic novel \\
train & 0 & a gingerbread steam train made of candy \\
train & 0 & a train in lava \\
bus & 0 & a black and white drawing of a bus \\
bus & 0 & a bus in cyberpunk style \\
lucia & 0 & an astronaut walking on mars \\
lucia & 0 & a claymation animation of a woman walking  \\
tennis & 15 & a robot throwing a laser ball \\
tennis & 15 & astronaut playing tennis on the surface of the moon \\
bear & 60 & a polar bear exploring on an iceberg \\
bear & 60 & a space bear walking beneath the stars  \\
flamingo & 0 & 2D vector animation of a group of flamingos standing near some rocks and water \\
flamingo & 0 & oil painting of pink flamingos wading \\
hike & 0 & a green alien explorer hiking in the mountains \\
hike & 0 & paper cut-out mountains with a paper cut-out hiker \\
goat & 59 & a tiger prowling along the ridge above a jungle \\
goat & 59 & a dragon prowling over a crater on the moon \\
parkour & 60 & a man jumping over rocks in a red sandstone canyon \\
parkour & 60 & a robot dodging through an obstacle course \\
cows & 10 & a pig standing in the mud \\
cows & 10 & a robotic cow walking along a muddy road \\
camel & 10 & a camel robot on a snowy day  \\
camel & 10 & toy camel standing on dirt near a fence \\
blackswan & 0 & a watercolor painting of a white swan \\
blackswan & 0 & a crochet black swan swims in a pond with rocks and vegetation \\
dog & 20 & a cat walking \\
dog & 20 & a dalmatian dog walking \\
kite-surf & 10 & a sand surfer kicking up sand in the desert \\
kite-surf & 10 & kite surfer in the ocean at sunset \\
libby & 0 & chinese ink painting of a dog running \\
libby & 0 & 3D animation of a small dog running  through grass \\
horsejump-high & 0 & a cartoon of a magical flying horse jumping over an obstacle \\
horsejump-high & 0 & person rides on a horse while jumping over an obstacle with an aurora borealis in the background \\

\bottomrule
\end{tabular}
\caption{\textbf{DAVIS stylization evaluation settings.}}
\end{table*}



\end{document}